\pdfoutput=1
\documentclass[lettersize,journal]{IEEEtran}
\usepackage[utf8]{inputenc}
\usepackage{amsmath,amsfonts}
\usepackage{algorithmic}
\usepackage{array}
\usepackage[caption=false,font=normalsize,labelfont=sf,textfont=sf]{subfig}
\usepackage{textcomp}
\usepackage{stfloats}
\usepackage{url}
\usepackage{verbatim}
\usepackage{graphicx}
\usepackage{cite}
\usepackage{booktabs}
\usepackage{xcolor}
\usepackage{tabularx}
\newcolumntype{Y}{>{\centering\arraybackslash}X}
\newcolumntype{S}{>{\hsize=0.3\hsize\centering\arraybackslash}X}
\newcolumntype{L}{>{\hsize=0.7\hsize\centering\arraybackslash}X}
\usepackage{multirow} 
\usepackage{arydshln}
\usepackage{algorithm,algorithmic}
\usepackage{hyperref}
\usepackage{doi}
\def\BibTeX{{\rm B\kern-.05em{\sc i\kern-.025em b}\kern-.08em
    T\kern-.1667em\lower.7ex\hbox{E}\kern-.125emX}}
\usepackage{balance}
\DeclareUnicodeCharacter{200B}{}
\DeclareUnicodeCharacter{202F}{\,}
\begin{document}
\title{DAMSDAN: Distribution-Aware Multi-Source Domain Adaptation Network for Cross-Domain EEG-based Emotion Recognition}
\author{Fo Hu, \IEEEmembership{Member, IEEE}, Can Wang, Qinxu Zheng, Xusheng Yang, \IEEEmembership{Member, IEEE}, Bin Zhou, Gang Li, Yu Sun, \IEEEmembership{Senior Member, IEEE}, Wen-an Zhang, \IEEEmembership{Senior Member, IEEE}
\thanks{This work was supported in part by the National Natural Science Foundation of China (62501528); in part by the Ningbo Science Innovation Yongjiang 2035 Key Technological Breakthrough Project (2024Z199); in part by the Natural Science Foundation of Zhejiang Province (No. LQ23F030015); in part by the National Key Research and Development Program of China (2022ZD0117902); in part by the National Natural Science Foundation of China (82172056); in part by the Zhejiang Provincial Natural Science Foundation of China (LR23F010003). (\textit{Corresponding author: Yu Sun and Wen-an Zhang.})}
\thanks{F. Hu is with the College of Information Engineering, Zhejiang University of Technology, Hangzhou  310023, China, and also with the Institute of Wenzhou, Zhejiang University, Wenzhou 325036, China.}
\thanks{C. Wang, Q. Zheng, X. Yang, and W. Zhang are with the College of Information Engineering, Zhejiang University of Technology, Hangzhou  310023, China, (e-mail: wazhang@zjut.edu.cn)}
\thanks{B. Zhou and G. Li are with the College of Mathematical Medicine, Zhejiang Normal University, Jinhua 321004, China.}
\thanks{Y. Sun is with the Key Laboratory for Biomedical Engineering of MOE of China, Department of Biomedical Engineering, Zhejiang University, Hangzhou 310027, China, and also with the Department of Colorectal Surgery and Oncology (Key Laboratory of Cancer Pevention and Intervention, MOE), The Second Affiliated Hospital, Zhejiang University School of Medicine, and also with the Institute of Wenzhou, Zhejiang University, Wenzhou 325006, China, and also with the College of Mathematical Medicine, Zhejiang Normal University, Jinhua 321004, China, (e-mail: yusun@zju.edu.cn).}}

\markboth{}
{DAMSDAN: Distribution-Aware Multi-Source Domain Adaptation Network for Cross-Domain EEG-based Emotion Recognition}

\maketitle

\begin{abstract}
Significant inter-individual physiological variability poses a critical challenge to the generalization capability of EEG-based emotion recognition models in cross-domain scenarios. Despite advances in domain adaptation, two major challenges remain in multi-source settings: (1) How to dynamically model the distributional heterogeneity across source domains and quantify their relevance to the target domain, thereby mitigating negative transfer; and (2) how to achieve fine-grained semantic consistency between source and target domains to strengthen the model's class-discriminative capacity. To this end, we introduce a novel deep learning framework named distribution-aware multi-source domain adaptation network (DAMSDAN) to address the above challenges. Specifically, DAMSDAN incorporates prototype-based constraints and adversarial training to guide feature encoder in learning discriminative and domain-invariant emotion representations. Furthermore, a domain-aware source weighting strategy based on maximum mean discrepancy is proposed to dynamically evaluate inter-domain shifts and reweight source contributions, effectively mitigating negative transfer effects. Besides, we propose a prototype-guided conditional alignment strategy, driven by dual pseudo-label interaction, which enhances pseudo-label reliability and leverages category-level prototype information to facilitate stable and fine-grained semantic alignment, thereby effectively alleviating pseudo-label noise propagation and semantic drift. The experiments conducted on the SEED and SEED-IV datasets reveal that DAMSDAN achieves average accuracies of 94.86\% and 79.78\% under cross-subject settings, and 95.12\% and 83.15\% under cross-session settings, respectively. On the large-scale FACED dataset, DAMSDAN further achieves a leading accuracy of 82.88\% under cross-subject settings. Extensive ablation studies and interpretability analyses corroborate the effectiveness of DAMSDAN in cross-domain EEG-based emotion recognition tasks.
\end{abstract}


\section{Introduction}
\label{section1}
\IEEEPARstart{A}{ffective} computing, which lies at the intersection of artificial intelligence and cognitive science, has demonstrated substantial potential in advancing the naturalness and emotional responsiveness of human-computer interaction. Its methodologies have been successfully applied in diverse domains, such as brain-computer interfaces~\cite{ref1}, virtual reality systems~\cite{ref2}, and neurorehabilitation~\cite{ref3}. Among the various modalities employed in affective computing, electroencephalogram (EEG) signals have garnered significant attention due to their noninvasive nature, millisecond-level temporal resolution, and robustness against intentional deception. Therefore, EEG-based emotion recognition is increasingly regarded as a promising approach for decoding nuanced emotional states. However, EEG signals inherently exhibit substantial inter-subject variability and nonstationarity, which present major challenges for conventional machine learning and deep learning methods to generalize effectively in cross-domain scenarios, such as cross-subject or cross-session emotion recognition. These challenges often lead to noticeable performance degradation when models are deployed on unseen subjects or sessions. To address these limitations, recent studies have increasingly employed transfer learning strategies to exploit labeled data from source domains (i.e., known subjects) and enhance model's adaptability to unlabeled target domains (i.e., unknown subjects). These approaches aim to minimize reliance on target annotations while mitigating performance degradation when recognizing emotions from EEG signals across domains. As a result, transfer learning has become a prominent and active research direction in affective computing~\cite{ref4}.

Current strategies for emotion recognition using EEG through transfer learning are commonly categorized into two paradigms: Domain generalization (DG) and domain adaptation (DA). Domain-invariant representations are learned in DG methods by extracting shared features across multiple source domains, thereby enabling generalization to new target domains without the need for training data from them. While theoretically promising, DG approaches often face difficulties in achieving stable and robust performance in real-world scenarios, especially when the source domain distribution does not sufficiently cover the diversity of target domains or when substantial inter-subject variability exists. In contrast, DA methods utilize unlabeled target domain data to align the feature distributions of source and target domains. By explicitly mitigating the domain shift caused by inter-subject variability, DA methods have shown greater practical effectiveness in improving generalization performance for EEG-based emotion recognition across domains, and are therefore regarded as a more viable solution for real-world applications. Most existing DA approaches adopt two principal alignment strategies: (1) Marginal distribution alignment, which seeks to learning domain-invariant features by aligning the overall feature distributions between the source and target domains; and (2) conditional distribution alignment, which focuses on reducing semantic discrepancies by aligning class-level feature distributions across domains. In practical applications, marginal distribution alignment emphasizes the learning of transferable features, whereas conditional distribution alignment focuses on enhancing class-level discriminability. However, consistency in marginal distributions does not necessarily guarantee semantic alignment across classes. Therefore, one of the key issues in cross-domain EEG-based emotion recognition lies in effectively balance domain invariance and class discriminability under semantic consistency constraints, while achieving joint alignment of both marginal and conditional distributions in a robust and reliable manner.

To mitigate the marginal distribution discrepancy across source as well as target domains, existing studies generally follow two technical paradigms: (1) Statistical criterion-based alignment and (2) adversarial training-based alignment. The former typically employs metrics such as maximum mean discrepancy (MMD) and correlation alignment to reduce distribution discrepancy in the feature space. Li et al.~\cite{ref5} employed MMD to align the marginal distributions between the source and target domains, and subsequently trained a classifier on the aligned shared feature space to enhance emotion recognition performance in the target domain. Ma et al.~\cite{ref6} proposed a multi-dimensional alignment strategy based on MMD, which jointly models frequency bands, spatial channels, and temporal dynamics, thereby enhancing the consistency and cross-subject transferability of emotional representations. While statistical approaches offer structural simplicity and computational efficiency, their performance tends to degrade under the inherently nonlinear and low signal-to-noise characteristics of EEG data, thereby resulting in unstable alignment in real-world scenarios. To overcome this limitation, adversarial training-based methods have been introduced. These methods leverage a domain discriminator equipped with a gradient reversal layer (GRL) to encourage the feature encoder to learn domain-invariant representations that are indistinguishable across different domains. Li et al.~\cite{ref7} introduced a cross-subject domain adaptation approach projecting EEG signals through hemisphere-referenced transformation and employing DANN for feature alignment to achieve robust emotion recognition. Zhou et al.~\cite{ref8} proposed a prototype-pair learning framework that combines convolutional neural network (CNN)-based feature extraction and adversarial alignment to suppress pseudo-label noise and improve emotion recognition performance. Despite the effectiveness of adversarial training in aligning marginal distributions, most existing approaches remain limited to the single-source domain adaptation (SSDA) paradigm, which assumes a fixed one-to-one correspondence between a single source domain and the target domain. In practice, target domain data frequently exhibit limited sample sizes and complex distributions. As a result, SSDA methods are unable to exploit complementary information from multiple source domains, which constrains their generalization performance. To address this limitation, researchers have proposed the multi-source domain adaptation (MSDA) framework~\cite{ref9,ref10,ref11}, which aims to enhance target domain modeling by leveraging diverse and complementary information from multiple source domains. However, many existing MSDA methods adopt an oversimplified assumption of distributional homogeneity among source domains, thereby overlooking the distributional heterogeneity and potential redundancy. Recent studies~\cite{ref12} have demonstrated that indiscriminately aggregating data from all source domains may introduce feature conflicts and information interference, thereby inducing negative transfer and degrading performance on the target domain. Therefore, a critical challenge in marginal distribution alignment under MSDA settings lies in dynamically modeling the relevance of each source to the target domain, with the goal of effectively exploiting informative sources while suppressing the adverse impact of irrelevant or noisy ones.

Moreover, aligning marginal distributions alone is insufficient to ensure semantic consistency across source and target domains in terms of categories.  Consequently, achieving fine-grained semantic alignment regarding conditional distributions has emerged as a critical challenge in EEG-based emotion recognition. Existing conditional distribution alignment methods can generally be classified into two categories: (1) Conditional adversarial alignment and (2) pseudo-label guided alignment. Conditional adversarial methods utilize both feature representations and their associated class predictions to form conditional inputs for the domain discriminator, thereby encouraging the feature encoder to learn representations that are simultaneously domain-invariant and class-discriminative. Long et al.~\cite{ref13} put forward a conditional adversarial approach that refines adversarial training using class prediction uncertainty, achieving precise semantic alignment and strong cross-subject generalization for emotion recognition using EEG signals. Ye et al.~\cite{ref14} proposed ADALAM, which incorporates hierarchical attention and class-conditioned adversarial learning to enhance feature discriminability and improve cross-domain generalization. Despite their promising performance, conditional adversarial methods often over-rely on source-domain labels, making them vulnerable to overfitting and semantic drift in unsupervised target domains. To overcome this limitation, pseudo-label-guided conditional alignment strategies have been introduced~\cite{ref15}. Li et al.~\cite{ref31} proposed a class-centroid contrastive alignment strategy that constructs class-conditional centroids using source labels and target pseudo-labels, and reinforces class separability through inter-class repulsion. Hong et al.~\cite{ref43} constructed class-specific prototypes and introduced contrastive and alignment losses to facilitate joint distribution alignment, demonstrating strong potential in improving conditional consistency. However, existing pseudo-labeling strategies suffer from two key limitations. First, they typically rely on single-perspective label generation without reliable validation, limiting the accuracy of class assignments. Second, conventional pseudo-label-based alignment methods lack stable semantic anchors, hindering the enforcement of consistent class-level semantics across domains. Therefore, it is critical to develop a robust pseudo-labeling strategy and incorporate structural information to enable fine-grained conditional alignment, thereby improving both the generalization and discriminative power of cross-domain EEG-based emotion recognition models.

A novel distribution-aware multi-source domain adaptation network (DAMSDAN) is proposed to address these challenges, which integrates a marginal distribution alignment (MDA) module and a conditional distribution alignment (CDA) module to achieve joint alignment between the source and target domains at both the global and class-specific semantic levels. Specifically, within the MDA module, a prototype consistency constraint (PCC) and an adversarial domain alignment (ADA) mechanism collaboratively guide the feature encoding (FE) module to learn emotion representations that are both domain-invariant and discriminative. In addition, a domain-aware source weighting (DASW) strategy is introduced to dynamically quantify the distribution divergence across each source-target domain pair, thereby adaptively reweighting their transfer contributions and mitigating negative transfer. In the CDA module, a dual pseudo-label collaboration (DPLC) strategy is introduced to enhance pseudo-label reliability, while a prototype-guided conditional alignment (PGCA) mechanism is employed to enforce class-level semantic consistency across domains. Collectively, these components enhance the model's generalization capability under both global and class-specific distribution shifts. Extensive ablation studies and comparative experiments are conducted on three benchmark EEG datasets for emotion recognition, namely SEED, SEED-IV, and FACED, which comprehensively demonstrate the effectiveness and robustness of the proposed DAMSDAN framework. This paper's key contributions can be summarized as follows:
\vspace{0.5em}
\begin{enumerate}
\item A novel MDA module is proposed, which integrates PCC, ADA, and MMD-based DASW to enhance domain-invariant feature learning and suppress negative transfer.
\vspace{0.5em}
\item We develop a CDA module that integrates DPLC and PGCA strategy, which jointly improve pseudo-label reliability and facilitate class-level semantic consistency between source and target domains.
\vspace{0.5em}
\item The proposed DAMSDAN model consistently outperforms state-of-the-art methods on SEED, SEED-IV, and FACED datasets, demonstrating superior generalization and robustness in cross-domain EEG emotion recognition.
\end{enumerate}
\section{Methods}
\label{section2}
In the MSDA setting for EEG-based emotion recognition, we consider $M$ labeled source domains $\left\{ {{S_m}} \right\}_{m = 1}^M$ and one unlabeled target domain $T$. All domains share a common feature space $X$ and label space $Y$, but exhibit significant distributional discrepancies. Specifically, each source domain ${S_m}$ consists of ${N_{{S_m}}}$ labeled EEG samples $\left\{ {{X_{{S_m}}},{Y_{{S_m}}}} \right\} = \left\{ {(x_i^{{S_m}},y_i^{{S_m}})} \right\}_{i = 1}^{{N_{{S_m}}}}$ , drawn from the joint distribution ${D_{{S_m}}} = \left\{ {{X_{{S_m}}},{Y_{{S_m}}}} \right\}$ over $X \times Y$. The target domain $T$ contains ${N_T}$ unlabeled samples $\left\{ {{X_T}} \right\} = \left\{ {x_j^T} \right\}_{j = 1}^{{N_T}}$, drawn from the marginal distribution ${D_T} = \left\{ {{X_T}} \right\}$ over $X$. The proposed MSDA approach aims to learn a mapping $G:{X_T} \to {Y_T}$ that enables accurate emotion recognition on the target domain. This is achieved by minimizing the discrepancy between the joint distributions of ${S_m}$ and $T$ in a shared latent feature space $Z$. The function $G$ represents a deep neural network, parameterized by the trainable parameters $\Theta $. For clarity and consistency in the subsequent derivations, all notations used throughout this paper can be found in Table~\ref{table_0}.

\begin{table}[t]
\centering
\caption{Notations and Descriptions}
\label{table_0}
\renewcommand\arraystretch{1.5}
\begin{tabularx}{0.7\linewidth}{S L} 
\specialrule{1pt}{0pt}{0pt}
\textbf{Notation} & \textbf{Description} \\
\hline
$S$ & source domain \\
$T$ & target domain \\
${D^{{S_m}}}$ & source dataset \\
${D^T}$ & target dataset \\
${N^{{S_m}}}$ & number of source samples \\
${N^T}$ & number of target samples \\
${x^{{S_m}}}$ & source feature \\
${x^T}$ & target feature \\
${y^{{S_m}}}$ & source label \\
${y^T}$ & target label \\
${\mu _c}$ & prototype features \\
${\mathop{\rm F}\nolimits}(\cdot)$ & feature encoding module \\
${\mathop{\rm C}\nolimits}(\cdot)$ & classifier module \\
${\mathop{\rm D}\nolimits}(\cdot)$ & domain discriminator module \\
\specialrule{1pt}{0pt}{0pt}
\end{tabularx}
\end{table}

\begin{figure*}[t!]\centering
\centering
\includegraphics[width=7in]{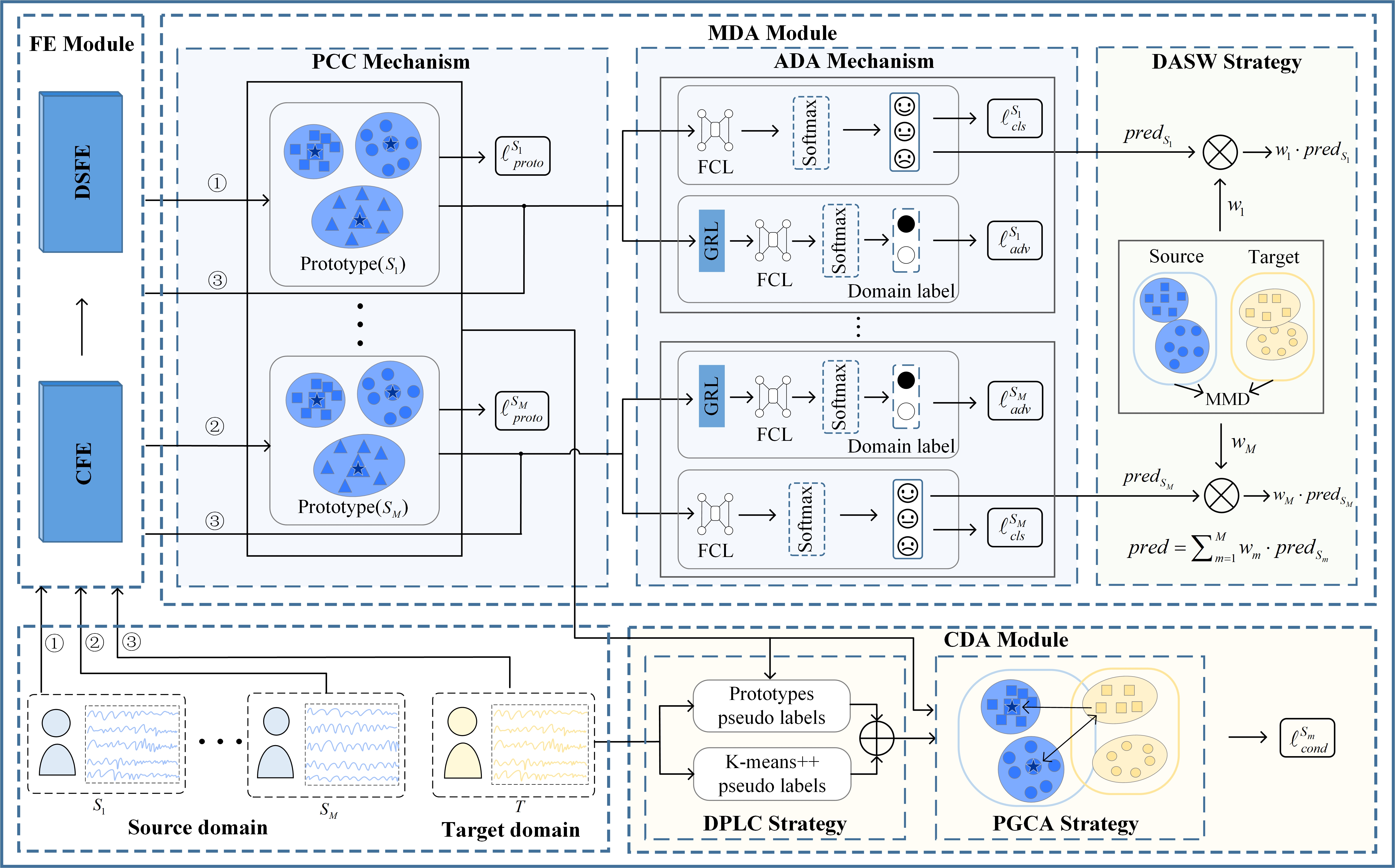}
\caption{Overall framework of the proposed DAMSDAN model.}
\label{fig_1}
\end{figure*}

The architecture of DAMSDAN, depicted in Fig.~\ref{fig_1}, is composed of three main components: The FE module, the MDA module, and the CDA module. The FE module integrates a common feature encoder (CFE) and a domain-specific feature encoder (DSFE) to extract transferable and discriminative emotion-related representations. The MDA module incorporates three components: The PCC mechanism, the ADA mechanism, and the DASW strategy. This module aligns marginal feature distributions by minimizing the adversarial loss ${\ell _{adv}}$ and enhances class-wise structure through the prototype loss ${\ell _{proto}}$, which enhances intra-class similarity and increases inter-class distinction. The DASW strategy dynamically assigning source-specific weights according to distributional divergence, thereby emphasizing informative sources and suppressing negative transfer effects. The CDA module includes the DPLC strategy and the PGCA strategy, both of which are designed to align conditional distributions. Specifically, DPLC enhances pseudo-label reliability in the target domain, while PGCA promotes semantic consistency across domains. Minimizing the conditional alignment loss ${\ell _{cond}}$ strengthens class-level alignment and contributes to more stable decision boundaries. All components are jointly optimized in an end-to-end manner by minimizing the multi-source classification loss ${\ell _{cls}}$ along with ${\ell _{adv}}$, ${\ell _{proto}}$, and ${\ell _{cond}}$, thereby simultaneously enhancing domain transferability and emotional discriminability. The overall optimization objective is defined as follows:
\begin{equation}
\ell  = {\ell _{cls}} - {\lambda _1}\cdot {\ell _{adv}} + {\lambda _2}\cdot {\ell _{proto}} + {\lambda _3}\cdot {\ell _{cond}}
\label{equ_1}
\end{equation}
where ${\lambda _1}$, ${\lambda _2}$, and ${\lambda _3}$ are trade-off hyperparameters controling the relative contribution among the loss components.

\subsection{Feature Encoding and Classifier Modules}

To learn features with both discriminative power and transferability, we propose the FE module, which consists of two components: The CFE and DSFE. The CFE projects samples from both source and target domains into a unified latent space to capture cross-domain invariant representations. Meanwhile, the DSFE incorporates a multi-head self-attention mechanism to capture global feature dependencies and extract informative inter-domain structural patterns, thereby facilitating more effective modeling of domain shifts. By combining shared and domain-specific representations, the FE module preserves transferable information while retaining domain-sensitive characteristics, thereby establishing a robust foundation for subsequent classification and distribution alignment.

\begin{figure}[t!]\centering
\centering
\includegraphics[width=3.5in]{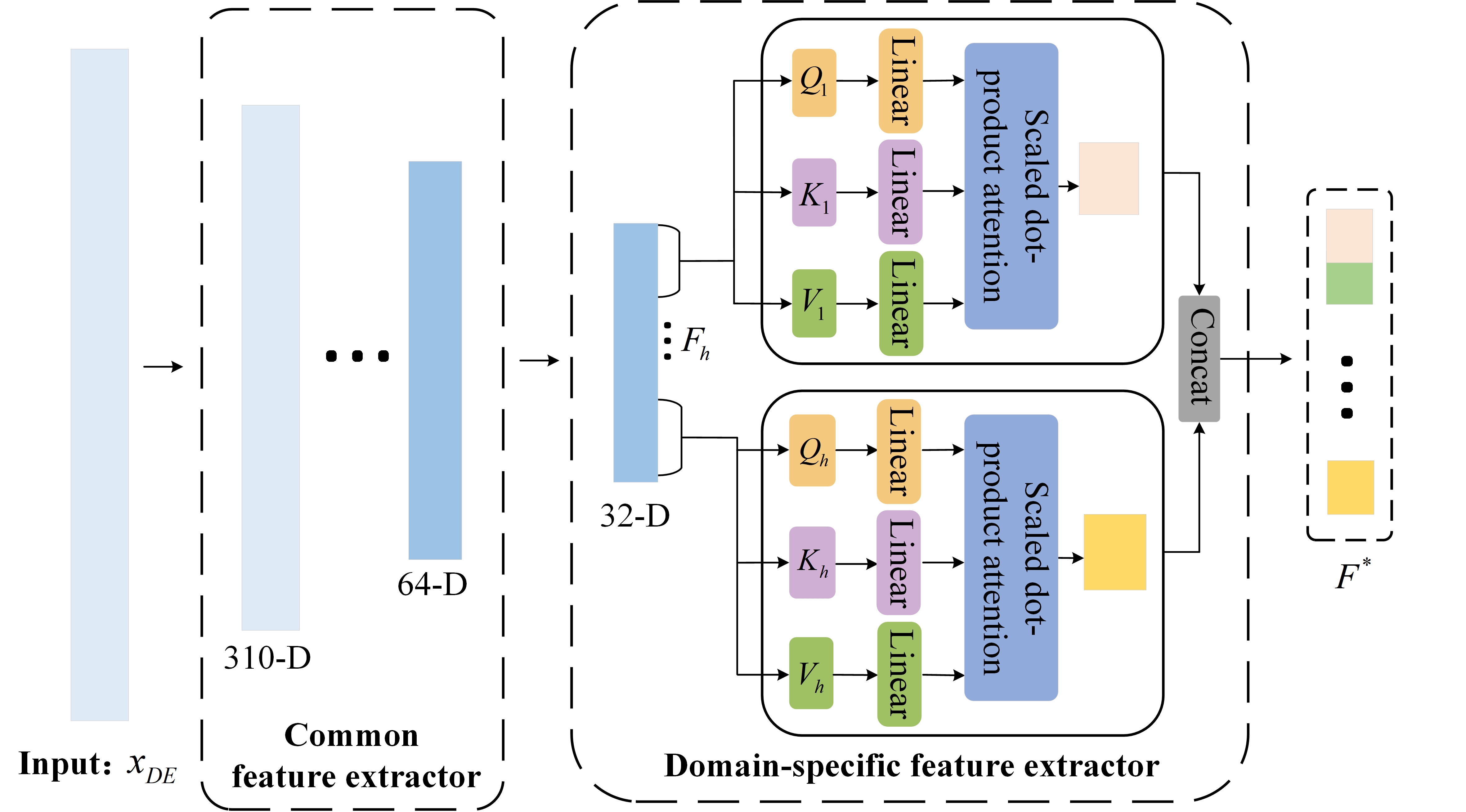}
\caption{Architecture of the FE module.}
\label{fig_2}
\end{figure}

Given the proven affective sensitivity and class discriminability of differential entropy (DE) features in EEG-based emotion recognition~\cite{ref18}, this paper employs DE features as model input, denoted as ${x_{DE}} \in {\mathbb{R}^{N \times D}}$, where $N$ represents the number of samples and $D$ denotes the feature dimension. The CFE consists of a fully connected network (FCN) with three layers containing 256, 128, and 64 neurons, respectively, performing progressive dimensionality reduction to extract domain-invariant latent representations. The DSFE comprises a single FCN layer with 32 neurons and incorporates a self-attention mechanism inspired by the Transformer architecture to capture long-range dependencies among input features, thereby facilitating the modeling of domain-specific structural variations. The overall network structure is illustrated in Fig.~\ref{fig_2}. All FCN layers adopt the LeakyReLU activation function with a negative slope coefficient of $\alpha  = 0.01$, which improves the model's sensitivity to negative-valued inputs. Batch normalization (BN) is subsequently applied to the FE module outputs to standardize feature distributions, thereby accelerating convergence and improving generalization during training. Given an input ${x_{DE}}$, the feature extraction process is formally defined as:
\begin{equation}
{f_{com}} = {{\mathop{\rm F}\nolimits} _{CFE}}\left( {{x_{DE}}} \right)
\label{equ_2}
\end{equation}
\begin{equation}
f_{spe}=\text{F}_{DSFE}\left( f_{com} \right) 
\label{equ_3}
\end{equation}
where ${{\mathop{\rm F}\nolimits} _{CFE}}\left( {\cdot} \right)$ and ${{\mathop{\rm F}\nolimits} _{DSFE}}\left( {\cdot} \right)$ denote the operations of the CFE and DSFE, respectively, and ${f_{com}} \in {\mathbb{R}^{N \times {D_c}}}$ and ${f_{spe}} \in {\mathbb{R}^{N \times {D_s}}}$ represent their corresponding output features.

Next, a multi-head self-attention mechanism is employed to capture latent global dependencies within the feature representation ${f_{spe}}$. Specifically, ${f_{spe}}$ is evenly partitioned into $H$ subspaces, producing a set of sub-feature matrices $\left\{ {{F_h}} \right\}_{h = 1}^H$, where ${F_h}\in{\mathbb{R}^{N \times {D_h}}}$ and ${D_h} = {{{D_s}} \mathord{\left/
 {\vphantom {{{D_s}} H}} \right.
 \kern-\nulldelimiterspace} H}$. For each sub-feature ${F_h}$, three learnable linear projection matrices ${\bf{W}}_h^Q \in {\mathbb{R}^{{D_h} \times {D_h}}}$, ${\bf{W}}_h^K \in {\mathbb{R}^{{D_h} \times {D_h}}}$, and ${\bf{W}}_h^V \in {\mathbb{R}^{{D_h} \times {D_h}}}$ are applied to generate the corresponding query ${Q_h}$, key ${K_h}$, and value ${V_h}$ matrices, respectively. The attention computation is defined as follows:
\begin{equation}
\left\{ {\begin{array}{*{20}{c}}
{{Q_h} = {F_h}\cdot {\bf{W}}_h^Q}\\
{{K_h} = {F_h}\cdot {\bf{W}}_h^K}\\
{{V_h} = {F_h}\cdot {\bf{W}}_h^V}
\end{array}} \right.
\label{equ_4}
\end{equation}

Subsequently, the scaled dot-product between the query and key matrices is calculated and passed through the Softmax function to yield attention weights, which are then used to weight the corresponding value matrix, producing the output of the $h$-th attention head, denoted as ${{SA}_h} \in {\mathbb{R}^{N \times {D_h}}}$. This process can be formally expressed as:
\begin{equation}
{{SA}_h} = {\mathop{\rm Softmax}\nolimits} \left( {\frac{{{Q_h}\cdot {K_h}^T}}{{\sqrt {{d_K}} }}} \right)\cdot {V_h}
\label{equ_5}
\end{equation}
where ${K_h}^T$ denotes the transpose of the key matrix ${K_h}$, and ${d_K}$ represents the dimensionality of ${K_h}$. Finally, the outputs of all attention heads are concatenated and linearly projected to recover the original dimensionality, resulting in the multi-head context-enhanced feature representation ${F^*} \in {\mathbb{R}^{N \times {D_s}}}$:
\begin{equation}
{F^*} = {\mathop{\rm Concat}\nolimits} \left( {{{SA}_1},{{SA}_2},...,{{SA}_H}} \right)\cdot {{\bf{W}}^*}
\label{equ_6}
\end{equation}
where ${\mathop{\rm Concat}\nolimits} \left(  \cdot  \right)$ denotes the concatenation function, and  ${{\bf{W}}^*}$ represents the learnable weight matrix used for linear projection.

In each source domain branch, the proposed classifier module comprises two FCNs responsible for predicting emotion categories of source domain samples. The second FCN employs the LeakyReLU activation function to enhance the model's capability for capturing nonlinear emotional representations. For the $m$-th source domain, the predicted labels $\hat Y_{}^{{S_m}}$ are obtained by passing the extracted features $F_{{S_m}}^*$ through the classifier. During training, the classification performance on the source domains is optimized using the cross-entropy loss, formally defined as:
\begin{equation}
{\ell _{cls}} = \frac{1}{M}\sum\limits_{m = 1}^M {\sum\limits_{i = 1}^{{N_{{S_m}}}} {{\mathop{\rm J}\nolimits} (\hat y_i^{{S_m}},y_i^{{S_m}})} } 
\label{equ_7}
\end{equation}
where $\hat y_i^{{S_m}}$ and $y_i^{{S_m}}$ denote the predicted and ground-truth labels of the $i$-th sample from the $m$-th source domain, respectively, and ${\mathop{\rm J}\nolimits} \left( {\cdot} \right)$ represents the standard cross-entropy loss function. The classification loss ${\ell _{cls}}$  encourages the model to learn discriminative EEG representations, thereby providing a reliable foundation for the subsequent distribution alignment modules.

\subsection{Marginal Distribution Alignment Module}
To achieve marginal distribution alignment of source and target domains within the feature space, the proposed MDA module comprises three key components: The PCC mechanism, the ADA mechanism, and the DASW strategy. The following outlines the details of each component:

{\bf PCC}: To improve the structural expressiveness of the extracted features in the semantic space, a prototype-based constraint is introduced. It is assumed that the feature representations corresponding to each emotional category  follow a Gaussian distribution $ N\left( {{\mu _c},\sigma _c^2} \right)$ in the latent embedding space. Accordingly, the prototype vector for category $c$, denote as ${\mu _c}$, is estimated as the mean of the feature representations assigned to that category. Given a labeled sample $\left\{ {(x_i^{{S_m}},y_i^{{S_m}})} \right\}_{i = 1}^{{N_{{S_m}}}}$ from the $m$-th  source domain and a feature encoder ${\mathop{\rm F}\nolimits} \left(  \cdot  \right)$, the class-wise prototype vector ${\mu _c}$ is computed as:
\begin{equation}
\mu _c^{{S_m}} = \frac{1}{{N_{{S_m}}^c}}\sum\limits_{i = 1}^{N_{{S_m}}^c} {{\mathop{\rm F}\nolimits} \left( {x_i^{{S_m}}} \right)} 
\label{equ_8}
\end{equation}
where $N_{{S_m}}^c$ denotes the sample count of category $c$ from the $m$-th source domain, and $\mu _c^{{S_m}}$ denotes the class prototype of category $c$ in source domain ${S_m}$, which is dynamically updated during training based on the evolving feature distribution. To promote class separability and enhance the structural expressiveness of the feature space, we introduce a prototype loss that minimizes the Euclidean distance of every sample from its respective class prototype. The loss function is defined as:
\begin{equation}
\begin{array}{l}
{\ell _{proto}} = \frac{1}{M}\sum\limits_{m = 1}^M {{E_{{S_m}}}\left[ { - \sum\limits_{{S_m}} {\log {P_{{S_m}}}\left( {{y^{{S_m}}} = c|{x^{{S_m}}}} \right)} } \right]} \\
\begin{array}{*{20}{c}}
{}&{}
\end{array} = \frac{1}{M}\sum\limits_{m = 1}^M {{E_{{S_m}}}\left[ { - \sum\limits_{{S_m}} {\log \frac{{\exp \left( { - {\mathop{\rm d}\nolimits} \left( {{\mathop{\rm F}\nolimits} \left( {{x^{{S_m}}}} \right),\mu _c^{{S_m}}} \right)} \right)}}{{\sum\nolimits_{\hat c = 1}^C {\exp \left( { - {\mathop{\rm d}\nolimits} \left( {{\mathop{\rm F}\nolimits} \left( {{x^{{S_m}}}} \right),\mu _{\hat c}^{{S_m}}} \right)} \right)} }}} } \right]} 
\end{array}
\label{equ_9}
\end{equation}
where ${P_{{S_m}}}\left( {{y^{{S_m}}} = c|{x^{{S_m}}}} \right)$ denotes the predicted probability that the sample ${x^{{S_m}}}$ belongs to class $c$, and ${\mathop{\rm d}\nolimits} \left( { \cdot , \cdot } \right)$ represents the Euclidean distance function.

{\bf ADA}: To match the global feature distributions of different domains, we utilize an ADA mechanism based on gradient reversal. Specifically, a domain discriminator is introduced to differentiate between source and target features, while the feature extractor ${\mathop{\rm F}\nolimits} \left(  \cdot  \right)$  is adversarially trained to produce domain-invariant representations. This is achieved through a GRL, which enables end-to-end adversarial learning. The domain adversarial loss is defined as:
\begin{equation}
\begin{array}{l}
{\ell _{adv}} = \frac{1}{M}\sum\limits_{m = 1}^M {\left( { - \sum\limits_{i = 1}^{{N_{{S_m}}}} {\log {\mathop{\rm D}\nolimits} \left( {{\mathop{\rm F}\nolimits} \left( {x_i^{{S_m}}} \right)} \right)} } \right.} \\
\left. {\begin{array}{*{20}{c}}
{}&{}
\end{array} - \sum\limits_{j = 1}^{{N_T}} {\log \left( {1 - {\mathop{\rm D}\nolimits} \left( {{\mathop{\rm F}\nolimits} \left( {x_j^T} \right)} \right)} \right)} } \right)
\end{array}
\label{equ_10}
\end{equation}

\begin{figure}[t!]\centering
\centering
\includegraphics[width=3.5in]{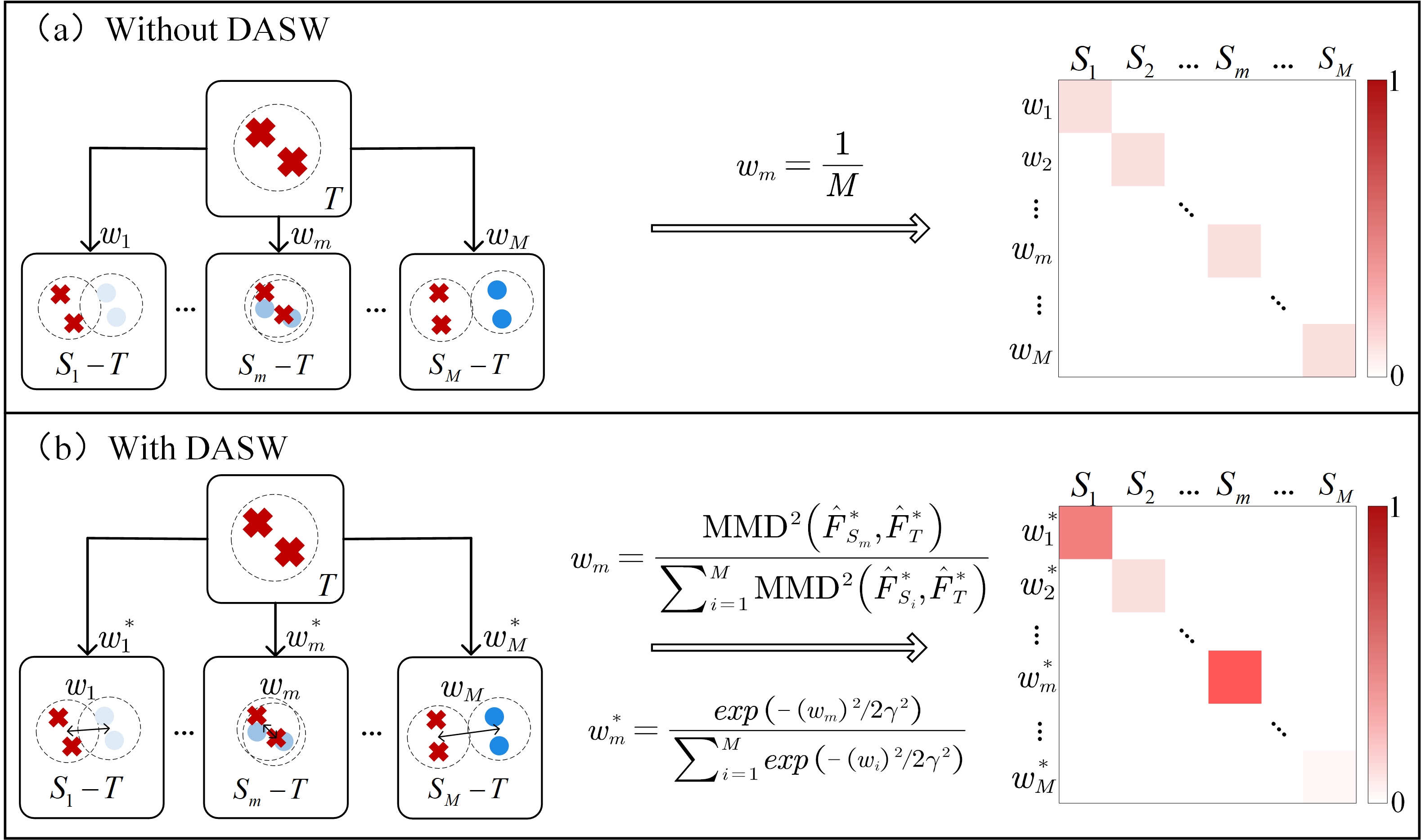}
\caption{Illustration of the DASW strategy.}
\label{fig_3}
\end{figure}

{\bf DASW}: To mitigate negative transfer arising from equal treatment of all source domains in multi-source adaptation, we propose the DASW that dynamically models the varying transfer contributions of each source domain to the target domain. As illustrated in Fig.~\ref{fig_3}, DASW strategy quantifies the feature distribution discrepancies between each source domain and the target domain, and assigns adaptive fusion weights accordingly. This approach amplifies the influence of semantically relevant sources while suppressing the interference of irrelevant or noisy ones, thereby improving the model's robustness and effectiveness.

Specifically, for the $m$-th source-target domain pair, the source domain features $\hat{F}_{S_m}^{*}$ and the target domain features $\hat{F}_{T}^{*}$, obtained after adversarial training, are formulated as follows:
\begin{equation}
\hat{F}_{S_m}^{*},\hat{F}_{T}^{*}=\mathrm{ADA}\left( F_{S_m}^{*},F_{T}^{*} \right) 
\label{equ_11}
\end{equation}
where $F_{{S_m}}^*$ and $F_T^*$ denote the source and target domain features prior to training, respectively. ${\mathop{\rm ADA}\nolimits} \left( {\cdot} \right)$ represents the adversarial domain alignment process. In order to assess transferability between source and target domains, the squared maximum mean discrepancy ${\rm{MM}}{{\rm{D}}^2}$ values are normalized across all source domains to derive the initial weight distribution:
\begin{equation}
\left\{ {\begin{array}{*{20}{l}}
{{w_m} = \frac{{{\rm{MM}}{{\rm{D}}^2}\left( {\hat F_{{S_m}}^*,\hat F_T^*} \right)}}{{\sum\nolimits_{i = 1}^M {{\rm{MM}}{{\rm{D}}^2}\left( {\hat F_{{S_i}}^*,\hat F_T^*} \right)} }}}\\
{{\rm{MM}}{{\rm{D}}^2}\left( {\hat F_{{S_m}}^*,\hat F_T^*} \right) = \frac{1}{{{N_{{S_m}}}^2}}{\rm{K}}\left( {\hat F_{{S_m}}^*,\hat F_{{S_m}}^*} \right)}\\
{\begin{array}{*{20}{c}}
{\begin{array}{*{20}{c}}
\begin{array}{l}
\begin{array}{*{20}{c}}
{}
\end{array}\begin{array}{*{20}{c}}
{}&{}&{}&{}&{}&{}
\end{array} + \frac{1}{{{N_T}^2}}{\rm{K}}\left( {\hat F_T^*,\hat F_T^*} \right)\\
\begin{array}{*{20}{c}}
{\begin{array}{*{20}{c}}
{}
\end{array}}&{}&{}&{}&{}&{}
\end{array} - \frac{2}{{{N_{{S_m}}} \cdot {N_T}}}{\rm{K}}\left( {\hat F_{{S_m}}^*,\hat F_T^*} \right)
\end{array}
\end{array}}
\end{array}}
\end{array}} \right.
\label{equ_12}
\end{equation}
where ${\mathop{\rm K}\nolimits} ( \cdot , \cdot )$ denotes a Gaussian kernel function.

To emphasize the contribution of source domains with lower distributional discrepancy, a Gaussian function is employed to further amplify their influence. The fusion weight for the $m$-th source domain is then computed via normalization as follows:
\begin{equation}
w_m^{*} = \frac{{exp\left( { - \frac{{{{\left( {{{w}_m}} \right)}^2}}}{{2{\gamma ^2}}}} \right)}}{{\sum\nolimits_{i = 1}^M {exp\left( { - \frac{{{{\left( {{{w}_i}} \right)}^2}}}{{2{\gamma ^2}}}} \right)} }}
\label{equ_13}
\end{equation}
where $\gamma $ controls the decay rate of the weighting function, thereby balancing the contribution between source domains with varying levels of correlation to the target domain.

During inference, the final prediction for a target domain sample $X_{}^T$ is computed via a weighted aggregation of the output probabilities from all source domain classifiers.
\begin{equation}
pred\left( {\hat F_T^*} \right) = \sum\limits_{m = 1}^M {{w_m^{*}} \cdot {{\rm{C}}_m}\left( {\hat F_T^*} \right)} 
\label{equ_14}
\end{equation}
where ${{{\rm{C}}_m}\left( {\hat F_T^*} \right)}$ denotes the class probability distribution predicted by the $m$-th source domain model for the target sample $X_{}^T$. In summary, the proposed DASW strategy adaptively adjusts the transfer contribution of each source domain model according to its actual distributional proximity to the target domain, thereby minimizing the risk of negative transfer in multi-source domain adaptation.

It is noteworthy that during the adversarial domain alignment process, the extracted features not only facilitate marginal distribution alignment via the domain discriminator, but also dynamically update the fusion weights in the DASW strategy (as defined in Eqs.~\eqref{equ_12} and \eqref{equ_13}). Specifically, at each training stage, the MMD distance between each source-target domain pair is computed based on the current feature representations, and the corresponding fusion weights are adaptively updated. This joint optimization fosters the synergy between marginal distribution alignment and source model fusion, thereby improving the overall effectiveness and stability of knowledge transfer.

\subsection{Conditional Distribution Alignment Module}

In unsupervised domain adaptation tasks, aligning only the marginal distributions between source and target domains and reducing the classification error on the source domain is often insufficient to achieve satisfactory transfer performance~\cite{ref19}. To enhance the model's generalization capability, it is essential to align conditional distributions by promoting feature consistency among samples of the same emotional class across different domains. To this end, we propose a CDA module that integrates two key strategies. First, the DPLC strategy is introduced to enhance the reliability of pseudo-labels within the target domain. Second, the PGCA strategy is used to construct and align class-level prototypes across domains, thereby mitigating semantic shift and improving cross-domain discriminability.

{\bf DPLC:} To reduce the negative effects of pseudo-label noise propagation on training stability, we propose a dual pseudo-label generation mechanism that exploits both discriminative and structural information. Specifically, the first type of pseudo-label is derived from classifier-predicted probabilities, reflecting the model's discriminative ability. The second type is obtained via the K-Means++ clustering algorithm, which captures the intrinsic structural distribution of target domain features within the latent space. To improve the quality of pseudo labels, a collaboration strategy is introduced, which selects high-confidence samples based on the consistency between two independently generated pseudo-label sources. These samples are subsequently incorporated into training to effectively suppress error accumulation and improve the robustness of pseudo-label learning.

The first type of pseudo-label is generated by assessing how similar a target sample is to the class prototypes in the source domains. Let $\mu _c^{{S_m}}$ denote the prototype of class $c$ in the $m$-th source domain. Then, for a target sample ${x^T}$, its probability of belonging to class $c$ is defined as:
\begin{equation}
{P_{{S_m}}}\left( {{y^T} = c|{x^T}} \right) = \frac{{\exp \left( { - {\mathop{\rm d}\nolimits} \left( {{\mathop{\rm F}\nolimits} \left( {{x^T}} \right),\mu _c^{{S_m}}} \right)} \right)}}{{\sum\nolimits_{i = 1}^C {\exp \left( { - {\mathop{\rm d}\nolimits} \left( {{\mathop{\rm F}\nolimits} \left( {{x^T}} \right),\mu _{i}^{{S_m}}} \right)} \right)} }}
\label{equ_15}
\end{equation}
where ${\mathop{\rm d}\nolimits} ( \cdot , \cdot )$ denotes the Euclidean distance function, ${\mathop{\rm F}\nolimits} ( \cdot )$ represents the feature encoder, and $C$ indicates the overall quantity of emotion classes. This method leverages semantic prototypes from the source domain to guide the class assignment for target samples, thereby generating pseudo-labels. However, owing to notable differences in data distributions across the source and target domains, directly relying on source-domain prototypes for label inference may lead to low pseudo-label confidence, increasing the risk of overfitting during target domain adaptation.

Therefore, a structure-aware pseudo-label generation strategy is proposed by applying the K-Means++ clustering algorithm to perform unsupervised clustering on target domain features. This method aims to uncover the latent structural organization of samples in the embedding space and produce pseudo-labels with improved structural consistency. Specifically, the source domain prototype $\mu _c^{{S_m}}$ for class $c$ used to initialize the corresponding cluster center $\mu _c^T$ in the target domain, thereby establishing a one-to-one semantic correspondence across categories. During training, $\mu _c^T$ is iteratively updated based on the clustering results, enabling progressive structural adaptation across training rounds. Based on the updated cluster centers, the conditional probability of assigning target sample ${x^T}$ to class $c$ is computed by evaluating its distances to all cluster centers. The probability is defined as follows:
\begin{equation}
{P_T}\left( {{y^T} = c|{x^T}} \right) = \frac{{\exp \left( { - {\mathop{\rm d}\nolimits} \left( {{\mathop{\rm F}\nolimits} \left( {{x^T}} \right),\mu _c^T} \right)} \right)}}{{\sum\nolimits_{i = 1}^C {\exp \left( { - {\mathop{\rm d}\nolimits} \left( {{\mathop{\rm F}\nolimits} \left( {{x^T}} \right),\mu _{i}^T} \right)} \right)} }}
\label{equ_16}
\end{equation}

Given that the two types of pseudo-labels are derived from the discriminative and structural perspectives, each capturing complementary aspects of the data, their consistency provides a critical criterion for evaluating pseudo-label reliability. Based on this insight, we introduce a pseudo-label collaboration learning mechanism to identify high-confidence target samples whose labels are consistent across both perspectives. These samples are then used to lead the training phase, facilitating the model to progressively focus on aligning distributions and enhancing class discrimination using the most reliable supervision. For a target sample ${x^T}$, let ${P_{{S_m}}}\left( {{y^T} = c|{x^T}} \right)$ and ${P_T}\left( {{y^T} = c|{x^T}} \right)$ denote the pseudo-label prediction probabilities from the discriminative and structural perspectives, respectively. The predicted class labels $\hat y_{S_m}^T$ and $\hat y_T^T$ are then obtained via the maximum probability criterion as follows:
\begin{equation}
\hat y_{S_m}^T = \mathop {\arg \max }\limits_{{y^T} \in Y} {P_{{S_m}}}\left( {{y^T} = c|{x^T}} \right)
\label{equ_17}
\end{equation}
\begin{equation}
\hat y_T^T = \mathop {\arg \max }\limits_{{y^T} \in Y} {P_T}\left( {{y^T} = c|{x^T}} \right)
\label{equ_18}
\end{equation}

Based on the prediction outputs, if the two kinds of pseudo-labels are consistent, i.e., $\hat y_{{S_m}}^T = \hat y_T^T$, the corresponding target sample is regarded as label-consistent and included in the consistency-based pseudo-labeled set ${D_{eq}} = \left\{ {{x^T}|\hat y_{{S_m}}^T = \hat y_T^T} \right\}$. To further enhance the reliability of pseudo-labels, we select the top ${{t{N^T}} \mathord{\left/
 {\vphantom {{t{N^T}} T}} \right.
 \kern-\nulldelimiterspace} T}$ samples from ${D_{eq}}$ with the highest prediction confidence at each training epoch $t$, where ${N^T}$ represents the overall quantity of target samples and $T$ represents the total quantity of training epochs. These selected samples are treated as high-confidence pseudo-supervised instances for the current training round. For each of these samples, the final pseudo-label is generated by fusing the predicted probabilities from both the discriminative and structural perspectives, with fusion weights dynamically adjusted according to the training progress. The fused pseudo-label for a target sample ${x^T}$ is computed as follows:
\begin{equation}
P\left( {{y^T}|{x^T}} \right) = \chi \cdot{P_T}\left( {{y^T} = c|{x^T}} \right) + \left( {1 - \chi } \right)\cdot{P_{{S_m}}}\left( {{y^T} = c|{x^T}} \right)
\label{equ_19}
\end{equation}
where $\chi  = {1 \mathord{\left/
 {\vphantom {1 {\left( {1 + {e^{ - k\left( {t - T/2} \right)}}} \right)}}} \right.
 \kern-\nulldelimiterspace} {\left( {1 + {e^{ - k\left( {t - T/2} \right)}}} \right)}}$. The final fused class label is assigned according to the maximum probability criterion, as defined below:
\begin{equation}
\hat y = \mathop {\arg \max }\limits_{{y^T}} P\left( {{y^T}|{x^T}} \right)
\label{equ_20}
\end{equation}

This fusion strategy effectively facilitates the model's gradual shift from source-domain supervision to target-domain pseudo-supervision. As training progresses, the model increasingly focuses on the intrinsic feature distribution of the target domain, thereby enhancing its ability to capture underlying structural patterns.

\begin{figure}[t!]\centering
\centering
\includegraphics[width=3.5in]{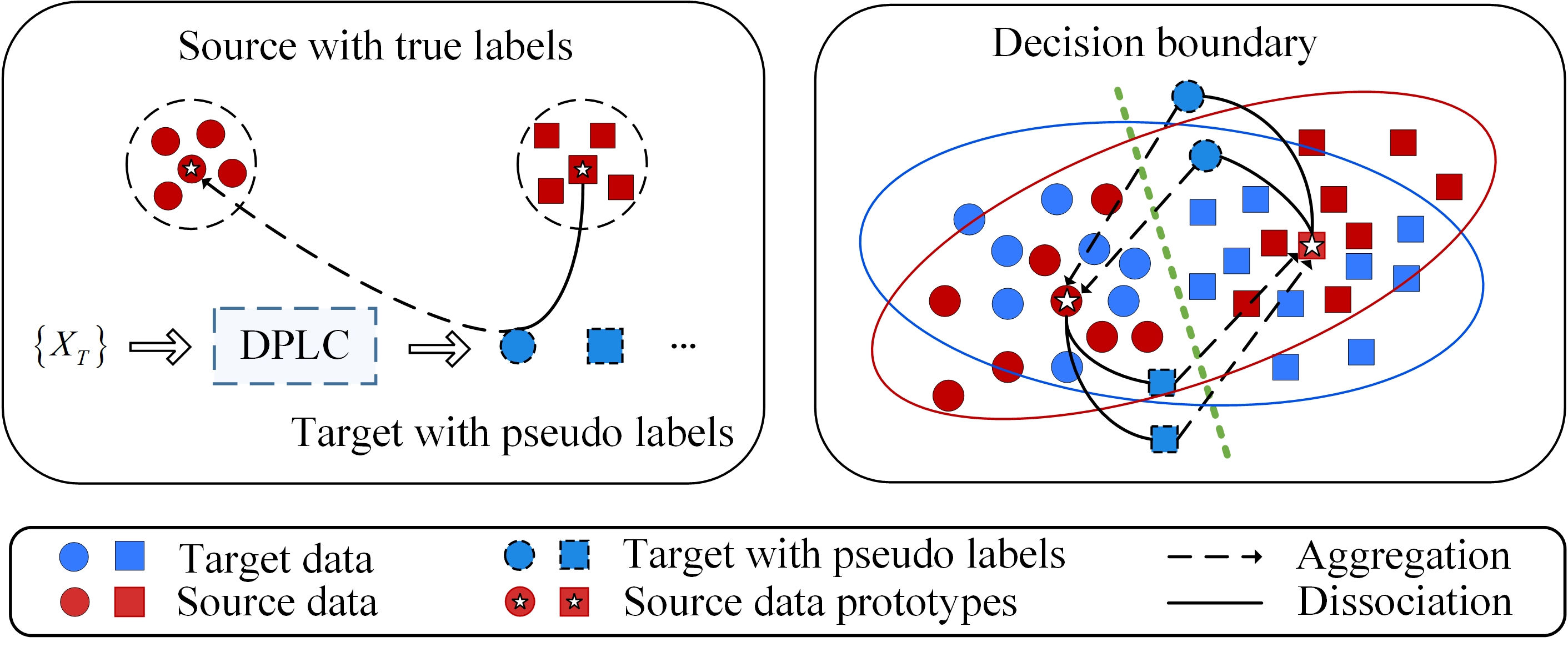}
\caption{Illustration of the PGCA strategy.}
\label{fig_4}
\end{figure}

{\bf PGCA:} To effectively mitigate conditional distribution shifts between source and target domains while preserving class-discriminative structures, we propose a prototype-guided conditional alignment strategy. As illustrated in Fig.~\ref{fig_4}, this method enhances class-level feature consistency by minimizing the conditional distribution discrepancy between target samples and source-domain class prototypes.

To further mitigate the conditional distribution shift between the source and target domains, this paper introduces a prototype-based conditional alignment loss. This loss quantitatively evaluates semantic divergence by computing the distance between source-domain class prototypes and corresponding target-domain features. The objective function is formally defined as follows:
\begin{equation}
\begin{array}{l}
{\ell _{cond}} = \frac{1}{M}\sum\limits_{m = 1}^M {\left( {\sum\limits_{{x^T} \in {X_T}} {\sum\limits_{c = 1}^C {1[{y^T} = c] \cdot {\mathop{\rm d}\nolimits} ({\mathop{\rm F}\nolimits} ({x^T}),\mu _c^{{S_m}})} } } \right.} \\
\left. {\begin{array}{*{20}{c}}
{}&{}
\end{array} - \beta \sum\limits_{{x^T} \in {X_T}} {\sum\limits_{c = 1}^C {1[{y^T} \ne c] \cdot {\mathop{\rm d}\nolimits} ({\mathop{\rm F}\nolimits} ({x^T}),\mu _c^{{S_m}})} } } \right)
\end{array}
\label{equ_21}
\end{equation}
where ${y^T}$ denotes the pseudo-label of the target sample (as defined in Eq.~\eqref{equ_20}), and $1[ \cdot ]$ corresponds to an indicator function that yields 1 if the condition is met, and 0 if not. This loss function aims to pull target features closer to their corresponding class prototypes when labels match, and push them apart otherwise. By promoting intra-class compactness and inter-class separability, it facilitates class-level conditional distribution alignment and effectively mitigates semantic shift across domains.

Finally, the proposed cross-domain emotion recognition model jointly optimizes four core loss components. The overall objective function is formulated as follows:
\begin{equation}
\ell  = {\ell _{cls}} - {\lambda _1}\cdot {\ell _{adv}} + {\lambda _2}\cdot {\ell _{proto}} + {\lambda _3}\cdot {\ell _{cond}}
\label{equ_21}
\end{equation}
where ${\ell _{cls}}$ denotes the source classification loss, ${\ell _{adv}}$ represents the adversarial domain alignment loss, ${\ell _{proto}}$ corresponds to the prototype consistency loss, and ${\ell _{cond}}$ refers to the conditional distribution alignment loss. The coefficients ${\lambda _1}$, ${\lambda _2}$, and ${\lambda _3}$ are trade-off parameters that balance the contributions of each loss component. By jointly incorporating discriminative enhancement, marginal distribution alignment, and conditional distribution alignment, the proposed objective function enables the learning of emotion representations that are both domain-generalizable and category-discriminative, thereby improving the robustness of cross-domain emotion recognition.

\section{Experimental Results}
\subsection{Datasets and Preprocessing}
To thoroughly assess the capability of the proposed DAMSDAN model in emotion recognition across domains based on EEG data, we perform comprehensive experiments on three publicly accessible benchmark EEG emotion datasets: SEED~\cite{ref20}, SEED-IV~\cite{ref21}, and FACED~\cite{ref22}. The core characteristics of each dataset, along with their respective preprocessing protocols, are summarized below.

SEED Dataset: This dataset, constructed by Zheng et al.~\cite{ref20}, is one of the most widely used benchmarks in EEG-based emotion recognition research. A 62-channel ESI Neuroscan system recorded EEG signals at a 1000 Hz sampling rate. The dataset comprises three emotion categories: Positive, negative, and neutral. Data were collected from 15 participants (8 females and 7 males, with a mean age of 23), each of whom watched 15 emotion-eliciting film clips, five for each category, across three recording sessions conducted on different days.

SEED-IV Dataset: This dataset serves as an expanded version of the SEED dataset, developed to support more fine-grained multi-class emotion recognition tasks. It adopts the same EEG acquisition protocol as SEED, utilizing a 62-channel ESI Neuroscan system. In contrast to SEED, the number of emotional categories is expanded from three to four: Happiness, neutral, sadness, and fear. During data collection, each participant viewed 24 emotion-inducing video clips, including six clips under each emotional label to induce the appropriate affective states.

FACED Dataset: This dataset, constructed by Zhang et al.~\cite{ref22}, is among the largest publicly available resources for EEG-based emotion recognition. It comprises 32-channel EEG recordings from 123 participants. During the experiment, each annotated with one of nine discrete emotional states: Pleasure, tenderness, inspiration, happiness, sadness, disgust, fear, anger, and neutrality. Following the protocol established in~\cite{ref38}, this study categorizes pleasure, inspiration, happiness, and tenderness as positive affective states, and anger, disgust, fear, and sadness as negative affective states, thereby formulating a binary emotion classification task.

Data Preprocessing: Downsampling to 200 Hz was first applied to the raw EEG signals. Then, a 0.3-50 Hz band-pass filter was then used to filter out low-frequency drifts and high-frequency interference, while preserving neural oscillatory components associated with emotional processing. To further attenuate residual artifacts, we applied independent component analysis. Subsequently, the continuous EEG signals were segmented into non-overlapping time windows of 1 second duration. DE features were then extracted from each window across five standard frequency bands: Delta (1-3 Hz), Theta (4-7 Hz), Alpha (8-13 Hz), Beta (14-30 Hz), and Gamma (31-50 Hz). In the SEED and SEED-IV datasets, each EEG segment comprises signals from 62 channels, resulting in a 310-element feature vector derived from 5 frequency bands across 62 channels. For the FACED dataset, which includes 32 EEG channels, each segment yields a 160-dimensional feature vector accordingly. To capture the temporal dynamics of emotional states and suppress high-frequency fluctuations unrelated to affective content, a linear dynamical system was applied to smooth the DE feature sequences. This process enhanced the temporal coherence of emotion-relevant patterns in the EEG data~\cite{ref8}.
\subsection{Implementation Details}
In this study, a multi-layer perceptron (MLP) architecture is adopted as the backbone to implement the feature encoder, classifier, and domain discriminator. All components utilize the LeakyReLU activation function, and network weights are initialized using a uniform distribution. As described previously, the FE module is composed of two submodules: The CFE and DSFE. The CFE architecture consists of: 310 (input layer)-256 (hidden layer \#1)-LeakyReLU-128 (hidden layer \#2)-LeakyReLU-64 (output layer). The DSFE is designed as: 64 (input layer)-LeakyReLU-32 (hidden layer)-LeakyReLU-4 head self-attention mechanism-32 (output layer). The classifier module comprises: 32 (input layer)-16 (hidden layer \#1)-LeakyReLU-$C$ (output layer). The domain discriminator follows: 32 (input layer)-16 (hidden layer)-LeakyReLU-1 (output layer)-Sigmoid activation.

During training, all parameters are optimized using the Adam optimizer with an initial learning rate of 2e-4 and a batch size of 256. All experiments were performed on a workstation featuring an NVIDIA GeForce RTX 4060 GPU, and the implementation is developed using the PyTorch framework with CUDA 10.0 support. Following training, we evaluate the computational complexity of DAMSDAN by measuring the average inference latency and the number of trainable parameters. The model yields an average inference time of 0.26 ms per EEG sample and consists of approximately 7.1 MB of trainable parameters. The complete code is hosted at: \url{https://github.com/ZJUTofBrainIntelligence/DAMSDAN}.

\subsection{Experimental Protocols}
To comprehensively evaluate the performance of the proposed DAMSDAN model in cross-domain emotion recognition tasks, we adopt two widely used validation protocols: (1) Cross-subject leave-one-subject-out cross-validation and (2) within-subject cross-session cross-validation. Specifically, for the SEED and SEED-IV datasets, both validation strategies are applied to evaluate the model's generalization across individuals and sessions. For the FACED dataset, which contains only single-session recordings per subject, only cross-subject leave-one-subject-out cross-validation is performed. The detailed experimental settings are as follows:

{\bf Cross-subject leave-one-subject-out cross-validation.} This strategy is among the most widely adopted evaluation protocols in EEG-based emotion recognition research~\cite{ref5,ref16,ref17,ref23,ref27}. In each iteration, data from one subject's session are designated as the unlabeled target domain, while data from the corresponding sessions of the remaining subjects serve as labeled source domains. Following prior works~\cite{ref8}, only the first session of each subject is used in this cross-subject evaluation. For the SEED and SEED-IV datasets, each subject provides a single session containing 3,394 EEG segments. Accordingly, the source domain comprises 14 subjects $\times$ session $\times$ 3,394 = 47,516 labeled samples, and the target domain includes 1 subject $\times$ 1 session $\times$ 3,394 = 3,394 unlabeled samples. Each subject is sequentially selected as the target domain, and the final performance is reported as the average recognition accuracy over all testing rounds. For the FACED dataset, which consists of 123 subjects with single-session recordings, we follow the evaluation protocol described in~\cite{ref38}: Where in 10 subjects are randomly selected as source domains, and the remaining 113 subjects are used individually as target domains. The final performance is presented as the mean recognition accuracy over all target subjects.

{\bf Within-subject cross-session cross-validation.} This evaluation protocol is employed to assess the model's temporal stability and its generalization capability across different recording sessions. Specifically, in both the SEED and SEED-IV datasets, each subject participates in three separate recording sessions conducted on different days. In each round of evaluation, the initial two sessions of a subject are considered the labeled source domain, with the final session acting as the unlabeled target domain. For each subject, the source domain contains 2 sessions $\times$ 3,394 samples = 6,788 samples, and the target domain contains 1 session $\times$ 3,394 samples = 3,394 samples. The session partitioning strategy for SEED-IV follows that of SEED. This procedure is repeated for all subjects, and the final performance is reported as the mean and standard deviation of classification accuracy across all target sessions.

We comprehensively compared the proposed DAMSDAN model with a wide spectrum of baseline methods, encompassing both traditional machine learning and deep learning-based domain adaptation methods. The traditional machine learning-based domain adaptation methods include: JDA~\cite{ref27}, MIMD~\cite{ref27}, MS-STM~\cite{ref27}, GFHF~\cite{ref27}, SA~\cite{ref27}, KNN~\cite{ref27}, PF~\cite{ref27}, CORAL~\cite{ref27}, GFK~\cite{ref24}, TKL~\cite{ref24}, TCA~\cite{ref25}, SVM~\cite{ref26}, TPT~\cite{ref26} and MEDA~\cite{ref28}. The deep learning-based domain adaptation methods comprise: PR-PL~\cite{ref8}, UDDA~\cite{ref31},  ADANN~\cite{ref43}, EPNNE~\cite{ref38}, DDC~\cite{ref27}, DAN~\cite{ref26}, MS-MDA~\cite{ref29}, MS-ADA~\cite{ref30}, DANN~\cite{ref32}, LRS~\cite{ref32}, MGFKD~\cite{ref33}, S2A2-MSDA~\cite{ref34}, ASJDA~\cite{ref35}, DGCNN~\cite{ref36}, DCCA~\cite{ref37}, MEERNET~\cite{ref39}, MSHCL~\cite{ref40}, CLISA~\cite{ref41} and MSMRA~\cite{ref42}.

\subsection{Experimental Results on Cross-Subject Leave-One-Subject-Out Cross-Validation}

We conducted cross-subject leave-one-subject-out cross-validation experiments on three publicly available EEG-based emotion recognition datasets: SEED, SEED-IV, and FACED. The performance comparisons between the proposed DAMSDAN model and a broad spectrum of state-of-the-art methods are presented in Tables~\ref{table_1}, \ref{table_2}, and~\ref{table_3}, respectively. On the SEED dataset, DAMSDAN achieved an average accuracy of 94.86\%$\pm $04.87\%, substantially outperforming all traditional machine learning and deep learning-based methods. Compared to the best-performing method PR-PL (93.06\%$\pm $05.12\%), DAMSDAN achieved a 1.8\% absolute improvement, underscoring its superior performance. For the more challenging SEED-IV dataset involving a four-class emotion classification task, DAMSDAN maintained strong performance with an average accuracy of 82.48\%$\pm $08.02\%, outperforming representative domain adaptation methods such as PR-PL, S2A2-MSDA, UDDA, and MS-CNN. On the large-scale FACED dataset, DAMSDAN achieved an accuracy of 82.88\%$\pm $07.77\%, once again ranking first among all compared methods. Except for the second-best method EPNNE, DAMSDAN consistently outperformed the remaining approaches by at least 5\%, demonstrating its robustness and scalability in large-sample scenarios. In summary, DAMSDAN consistently attains leading performance on all three benchmark EEG emotion datasets, confirming its effectiveness and generalization capability in cross-subject emotion recognition.

\begin{figure}[t!]\centering
\centering
\includegraphics[width=3.5in]{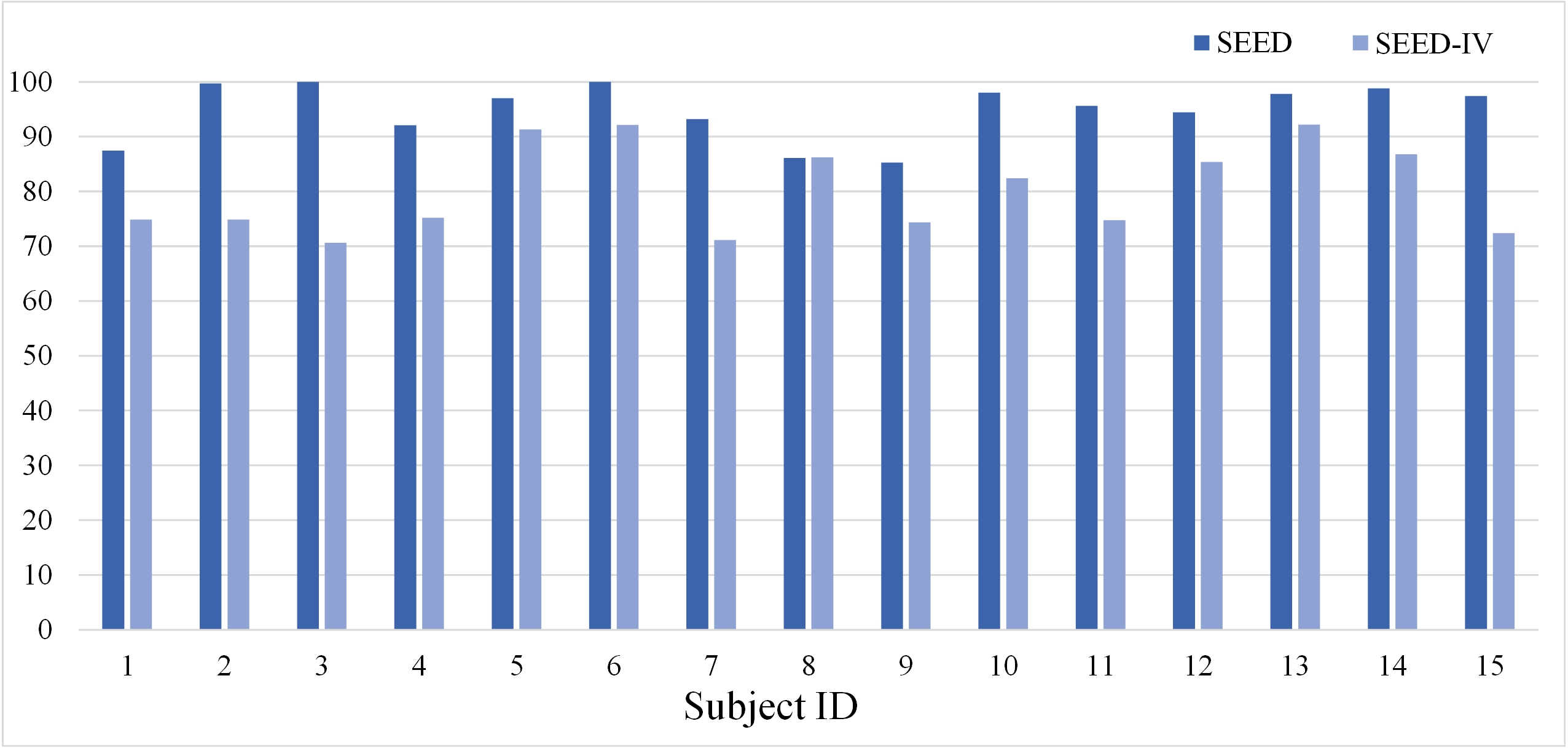}
\caption{Subject-wise classification accuracy under cross-subject leave-one-subject-out cross-validation on the SEED and SEED-IV datasets.}
\label{fig_5}
\end{figure}

To further assess the subject-wise adaptability of the proposed DAMSDAN model, we analyze the classification performance for each individual under the cross-subject leave-one-subject-out cross-validation scheme on the SEED and SEED-IV datasets. As illustrated in Fig.~\ref{fig_5}, DAMSDAN achieves over 85\% accuracy for all 15 subjects in the SEED dataset, with more than half exceeding 95\%. Notably, the DAMSDAN model achieves near-perfect performance for Subjects 2, 3, and 6, with accuracies of 99.71\%, 100\%, and 100\%, respectively. In the SEED-IV dataset, while subject-wise variability is more pronounced, most subjects attain accuracies between 70\% and 90\%, with Subjects 5, 6, and 13 achieving the highest accuracies of 91.23\%, 92.18\%, and 92.20\%, respectively. These results demonstrate that DAMSDAN consistently yields strong emotion recognition performance across subjects.
\begin{table}[t]
\centering
\caption{Performance of different models on the SEED dataset using cross-subject leave-one-subject-out cross-validation.}
\label{table_1}
\renewcommand{\arraystretch}{1.3}
\setlength{\tabcolsep}{4pt} 
\begin{tabular}{p{2.4cm}r|p{2.4cm}r}
\specialrule{1pt}{0pt}{0pt}
\textbf{Method} & ${P_{acc}}(\%)$ & \textbf{Method} & ${P_{acc}}(\%)$ \\
\cline{1-2} \cline{3-4}
\multicolumn{4}{c}{\textbf{Traditional Machine Learning Methods}} \\
\cline{1-2} \cline{3-4}
GFK~\cite{ref24} & 71.31$\pm$14.09 & TCA~\cite{ref25} & 63.64$\pm$14.88 \\
TKL~\cite{ref24} & 63.54$\pm$15.47 & SVM~\cite{ref26} & 58.18$\pm$13.85 \\
TPT~\cite{ref26} & 75.17$\pm$12.83 & JDA~\cite{ref27} & 73.50$\pm$06.30 \\
\cline{1-2} \cline{3-4}
\multicolumn{4}{c}{\textbf{Deep Learning Methods}} \\
\cline{1-2} \cline{3-4}
MS-MDA~\cite{ref29} & 81.43$\pm$10.17 & JDA-Net~\cite{ref32} & 86.70$\pm$11.39 \\
UDDA~\cite{ref31} & 88.10$\pm$06.54 & MGFKD~\cite{ref33} & 90.21$\pm$07.57 \\
EPNNE~\cite{ref38} & 89.10$\pm$03.60 & S2A2-MSDA~\cite{ref34} & 90.11$\pm$07.32 \\
PR-PL*~\cite{ref8} & 93.06$\pm$05.12 & ASJDA~\cite{ref35} & 88.70$\pm$13.54 \\
\cline{1-2} \cline{3-4}
\textbf{DAMSDAN} & \multicolumn{2}{r}{~} & \textbf{94.86$\pm$04.87} \\
\specialrule{1pt}{0pt}{0pt}
\end{tabular}
\vspace{-2mm}
\begin{flushleft}
\scriptsize * indicates the second-best model.
\end{flushleft}
\end{table}

\begin{table}[t]
\centering
\caption{Performance of different models on the SEED-IV dataset using cross-subject leave-one-subject-out cross-validation.}
\label{table_2}
\renewcommand{\arraystretch}{1.3}
\setlength{\tabcolsep}{4pt} 
\begin{tabular}{p{2.4cm}r|p{2.4cm}r}
\specialrule{1pt}{0pt}{0pt}
\textbf{Method} & ${P_{acc}}(\%)$ & \textbf{Method} & ${P_{acc}}(\%)$ \\
\cline{1-2} \cline{3-4}
\multicolumn{4}{c}{\textbf{Traditional Machine Learning Methods}} \\
\cline{1-2} \cline{3-4}
GFK~\cite{ref24} & 44.04$\pm$09.31 & TCA~\cite{ref25} & 37.01$\pm$10.47 \\
MIMD~\cite{ref27} & 60.22$\pm$08.69 & MS-STM~\cite{ref27} & 61.41$\pm$09.72 \\
GFHF~\cite{ref27} & 49.29$\pm$07.60 & JDA~\cite{ref27} & 54.60$\pm$05.30 \\
\cline{1-2} \cline{3-4}
\multicolumn{4}{c}{\textbf{Deep Learning Methods}} \\
\cline{1-2} \cline{3-4}
MS-MDA~\cite{ref29} & 59.34$\pm$05.48 & MS-ADA~\cite{ref30} & 59.29$\pm$13.65 \\
UDDA~\cite{ref31} & 73.14$\pm$09.43 & DAN~\cite{ref26} & 58.89$\pm$08.13 \\
S2A2-MSDA~\cite{ref34} & 76.23$\pm$09.02 & DANN~\cite{ref32} & 54.63$\pm$08.03 \\
PR-PL*~\cite{ref8} & 81.32$\pm$08.53 & ASJDA~\cite{ref35} & 79.58$\pm$17.19 \\
\cline{1-2} \cline{3-4}
\textbf{DAMSDAN} & \multicolumn{2}{r}{~} & \textbf{82.48$\pm$08.02} \\
\specialrule{1pt}{0pt}{0pt}
\end{tabular}
\vspace{-2mm}
\begin{flushleft}
\scriptsize * indicates the second-best model.
\end{flushleft}
\end{table}

\subsection{Experimental Results of Within-Subject Cross-Session Cross-Validation}
\begin{table}[t]
\centering
\caption{Performance of different models on the FACED dataset using cross-subject leave-one-subject-out cross-validation.}
\label{table_3}
\renewcommand{\arraystretch}{1.3}
\setlength{\tabcolsep}{4pt} 
\begin{tabular}{p{2.4cm}r|p{2.4cm}r}
\specialrule{1pt}{0pt}{0pt}
\textbf{Method} & ${P_{acc}}(\%)$ & \textbf{Method} & ${P_{acc}}(\%)$ \\
\cline{1-2} \cline{3-4}
\multicolumn{4}{c}{\textbf{Traditional Machine Learning Methods}} \\
\cline{1-2} \cline{3-4}
GFK~\cite{ref24} & 60.40$\pm$08.08 & TCA~\cite{ref25} & 56.35$\pm$09.68 \\
JDA~\cite{ref27} & 61.54$\pm$07.45 & MEDA~\cite{ref28} & 62.46$\pm$08.71 \\
\cline{1-2} \cline{3-4}
\multicolumn{4}{c}{\textbf{Deep Learning Methods}} \\
\cline{1-2} \cline{3-4}
LRS~\cite{ref32} & 65.83$\pm$07.93 & DGCNN~\cite{ref36} & 68.25$\pm$07.92 \\
DCCA~\cite{ref37} & 74.64$\pm$06.07 & EPNNE*~\cite{ref38} & 82.81$\pm$03.28 \\
MSHCL~\cite{ref40} & 77.00$\pm$03.60 & CLISA~\cite{ref41} & 75.10$\pm$03.50 \\
\cline{1-2} \cline{3-4}
\textbf{DAMSDAN} & \multicolumn{2}{r}{~} & \textbf{82.88$\pm$07.71} \\
\specialrule{1pt}{0pt}{0pt}
\end{tabular}
\vspace{-2mm}
\begin{flushleft}
\scriptsize * indicates the second-best model.
\end{flushleft}
\end{table}

\begin{figure}[t!]\centering
\centering
\includegraphics[width=3.5in]{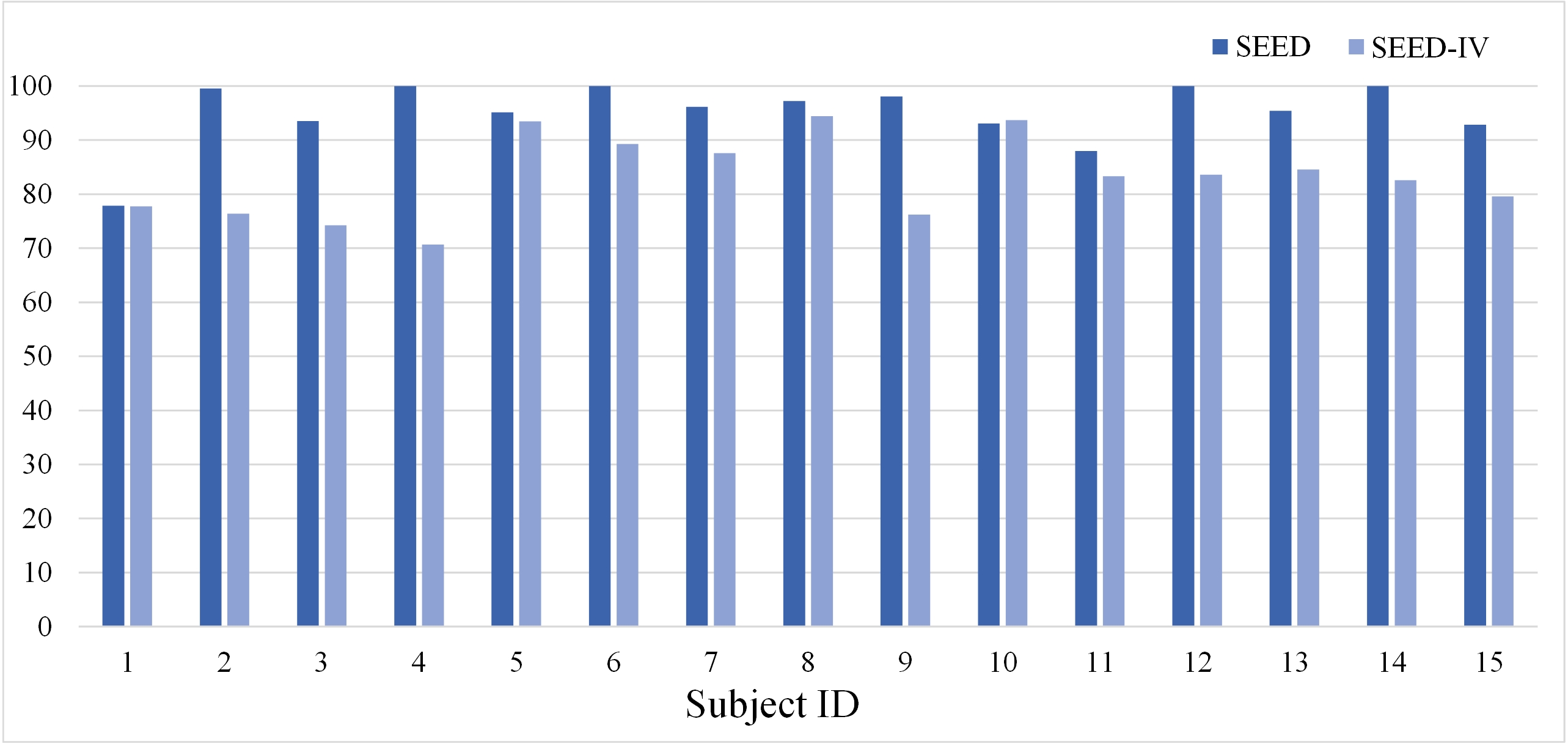}
\caption{Subject-wise classification accuracy under within-subject cross-session cross-validation on the SEED and SEED-IV datasets.}
\label{fig_6}
\end{figure}
\begin{table}[t]
\centering
\caption{Performance of different models on the SEED dataset using within-subject cross-session cross-validation.}
\label{table_4}
\renewcommand{\arraystretch}{1.3}
\setlength{\tabcolsep}{4pt} 
\begin{tabular}{p{2.4cm}r|p{2.4cm}r}
\specialrule{1pt}{0pt}{0pt}
\textbf{Method} & ${P_{acc}}(\%)$ & \textbf{Method} & ${P_{acc}}(\%)$ \\
\cline{1-2} \cline{3-4}
\multicolumn{4}{c}{\textbf{Traditional Machine Learning Methods}} \\
\cline{1-2} \cline{3-4}
SA~\cite{ref27} & 69.84$\pm$09.46 & KNN~\cite{ref27} & 75.68$\pm$13.82 \\
GFK~\cite{ref24} & 78.79$\pm$09.39 & TCA~\cite{ref25} & 74.27$\pm$12.88 \\
PF~\cite{ref27}  & 76.42$\pm$11.15 & CORAL~\cite{ref27} & 84.18$\pm$09.81 \\
\cline{1-2} \cline{3-4}
\multicolumn{4}{c}{\textbf{Deep Learning Methods}} \\
\cline{1-2} \cline{3-4}
MS-MDA~\cite{ref29} & 89.63$\pm$06.79 & DAN~\cite{ref26} & 80.00$\pm$08.88 \\
UDDA~\cite{ref31} & 90.19$\pm$10.07 & DANN~\cite{ref32} & 84.45$\pm$07.92 \\
DDC~\cite{ref27} & 81.53$\pm$06.83 & MSMRA~\cite{ref42} & 88.31$\pm$06.23 \\
PR-PL*~\cite{ref8} & 93.18$\pm$06.55 & ADANN~\cite{ref43} & 90.03$\pm$06.99 \\
\cline{1-2} \cline{3-4}
\textbf{DAMSDAN} & \multicolumn{2}{r}{~} & \textbf{95.12$\pm$05.92} \\
\specialrule{1pt}{0pt}{0pt}
\end{tabular}
\vspace{-2mm}
\begin{flushleft}
\scriptsize * indicates the second-best model.
\end{flushleft}
\end{table}

\begin{table}[t]
\centering
\caption{Performance of different models on the SEED-IV using within-subject cross-session cross-validation.}
\label{table_5}
\renewcommand{\arraystretch}{1.3}
\setlength{\tabcolsep}{4pt} 
\begin{tabular}{p{2.4cm}r|p{2.4cm}r}
\specialrule{1pt}{0pt}{0pt}
\textbf{Method} & ${P_{acc}}(\%)$ & \textbf{Method} & ${P_{acc}}(\%)$ \\
\cline{1-2} \cline{3-4}
\multicolumn{4}{c}{\textbf{Traditional Machine Learning Methods}} \\
\cline{1-2} \cline{3-4}
SA~\cite{ref27} & 52.81$\pm$16.28 & KNN~\cite{ref27} & 54.18$\pm$16.28 \\
GFK~\cite{ref24} & 56.14$\pm$15.13 & TCA~\cite{ref25} & 51.88$\pm$15.84 \\
PF~\cite{ref27}  & 60.27$\pm$16.36 & CORAL~\cite{ref27} & 66.06$\pm$15.13 \\
\cline{1-2} \cline{3-4}
\multicolumn{4}{c}{\textbf{Deep Learning Methods}} \\
\cline{1-2} \cline{3-4}
MS-MDA~\cite{ref29} & 61.43$\pm$15.71 & DAN~\cite{ref26} & 48.39$\pm$06.97 \\
MEERNet~\cite{ref39} & 72.10$\pm$14.10 & DANN~\cite{ref32} & 47.66$\pm$08.38 \\
PR-PL~\cite{ref8} & 74.62$\pm$14.15 & MSMRA~\cite{ref42} & 72.38$\pm$10.12 \\
ADANN~\cite{ref43} & 77.79$\pm$13.97 & S2A2-MSDA*~\cite{ref34} & 80.07$\pm$12.13 \\
\cline{1-2} \cline{3-4}
\textbf{DAMSDAN} & \multicolumn{2}{r}{~} & \textbf{83.15$\pm$07.43} \\
\specialrule{1pt}{0pt}{0pt}
\end{tabular}
\vspace{-2mm}
\begin{flushleft}
\scriptsize * indicates the second-best model.
\end{flushleft}
\end{table}

To further assess the recognition performance and generalization capability of the proposed DAMSDAN model under temporal distribution shift, we conducted within-subject cross-session cross-validation experiments on the SEED and SEED-IV datasets. The results are presented in Table~\ref{table_4} and Table~\ref{table_5}, respectively. On the SEED dataset, DAMSDAN achieved an average accuracy of 95.12\%$\pm $5.92\%, substantially outperforming all baseline methods. Notably, compared to the second-best deep domain adaptation method PR-PL (93.18\%$\pm $06.55\%), DAMSDAN achieved an absolute improvement of 1.94\%. Furthermore, on the SEED-IV dataset, DAMSDAN attained the highest accuracy of 83.15\%$\pm $07.43\%, significantly surpassing both the representative deep learning method S2A2-MSDA (80.07\%$\pm $12.13\%) and the traditional machine learning method CORAL (66.06\%$\pm $15.13\%). These results demonstrate that DAMSDAN not only maintains high classification accuracy across sessions but also exhibits strong robustness and adaptability to temporal distribution variations.

To further evaluate the cross-session generalization capability of the proposed DAMSDAN model at the individual level, we report the session-independent classification accuracy for each subject, as illustrated in Fig.~\ref{fig_6}. Overall, DAMSDAN exhibits strong generalization performance on the SEED dataset, achieving accuracies above 90\% for nearly all subjects. Remarkably, the DAMSDAN model attains near-perfect accuracies of 99.99\%, 99.98\%, 100\%, and 99.79\% for subiects 4, 6, 12 and 14, respectively. In contrast, although the overall performance slightly decreases on the more challenging SEED-IV dataset, DAMSDAN still maintains consistent and robust results across most subjects. In particular, subiects 5, 8, and 10 achieveaccuracies notably above the average. These results further substantiate the subject-level stability and effectiveness of DAMSDAN in cross-session EEG-based emotion recognition.

\section{Discussion}
In this section, we perform parameter sensitivity analyses on the SEED dataset using the cross-subject leave-one-subject-out cross-validation protocol. The objective is to systematically investigate the influence of key modules and hyperparameter settings on model performance, thereby elucidating the contribution of each component to the overall effectiveness in EEG-based emotion recognition.

\begin{table}[t]
\centering
\caption{Performance comparison in the ablation study.}
\label{table_6}
\renewcommand{\arraystretch}{1.3}
\setlength{\tabcolsep}{6pt} 
\begin{tabular}{p{4.2cm}r}
\specialrule{1pt}{0pt}{0pt}
\textbf{Ablation study} & ${P_{acc}}(\%)$ \\
\hline
w/o ADA mechanism & 86.40$\pm$06.04 \\
w/o PCC mechanism & 94.16$\pm$04.96 \\
w/o DASW strategy & 92.66$\pm$06.04 \\
w/o CDA module           & 92.16$\pm$05.14 \\
w/o DPLC strategy  & 93.10$\pm$05.08 \\
\hline
\textbf{DAMSDAN} & \textbf{94.86$\pm$04.87} \\
\specialrule{1pt}{0pt}{0pt}
\end{tabular}
\end{table}
\subsection{Ablation Study}

To systematically evaluate the individual contribution of each core module within the DAMSDAN model, a series of ablation experiments were conducted on the SEED dataset. All experiments were performed under identical training configurations, with only the specific component under investigation removed or modified in each case. The results are summarized in Table~\ref{table_6}, and detailed analyses are provided below.

{\bf ADA mechanism:} To assess the importance of the ADA mechanism in marginal distribution alignment, we removed the domain discriminator from the original model to construct a baseline variant. Under identical training conditions, the recognition accuracy decreased markedly from 94.86\% to 86.40\%, indicating a performance degradation of 8.46 percentage points.

{\bf PCC mechanism:} To evaluate the impact of the PCC mechanism, this module was removed from the full model while retaining the adversarial training component. Experimental results show that the classification accuracy dropped slightly to 94.16\%.

{\bf DASW Strategy:} To verify the effectiveness of the DASW strategy in multi-source domain adaptation, this module was removed from the full model for comparative analysis. The results show a notable decline in recognition accuracy from 94.86\% (full model) to 92.66\%, corresponding to a performance degradation of 2.20 percentage points.

{\bf CDA module:} To assess the performance of the CDA module, we conducted experiments under two settings: (1) Removing the entire CDA module; and (2) replacing the dual pseudo-label collaboration strategy with a conventional single pseudo-label mechanism~\cite{ref31}. The former led to a performance decline to 92.16\%, while the latter yielded an accuracy of 93.10\%. 

In summary, the ablation study results clearly demonstrate that each core component of the proposed DAMSDAN model, including ADA mechanism, PCC mechanism, DASW strategy, CDA module, and DPLC strategy, contributes substantially to the overall performance improvement. Among these, the ADA mechanism proves to be the most critical, as its removal leads to the most pronounced performance degradation. In contrast, the PCC mechanism has the least impact, though it still contributes meaningfully to improving feature discriminability. These findings validate the architectural design and confirm the effectiveness of the proposed framework.

\begin{figure}[t!]\centering
\centering
\includegraphics[width=3.4in]{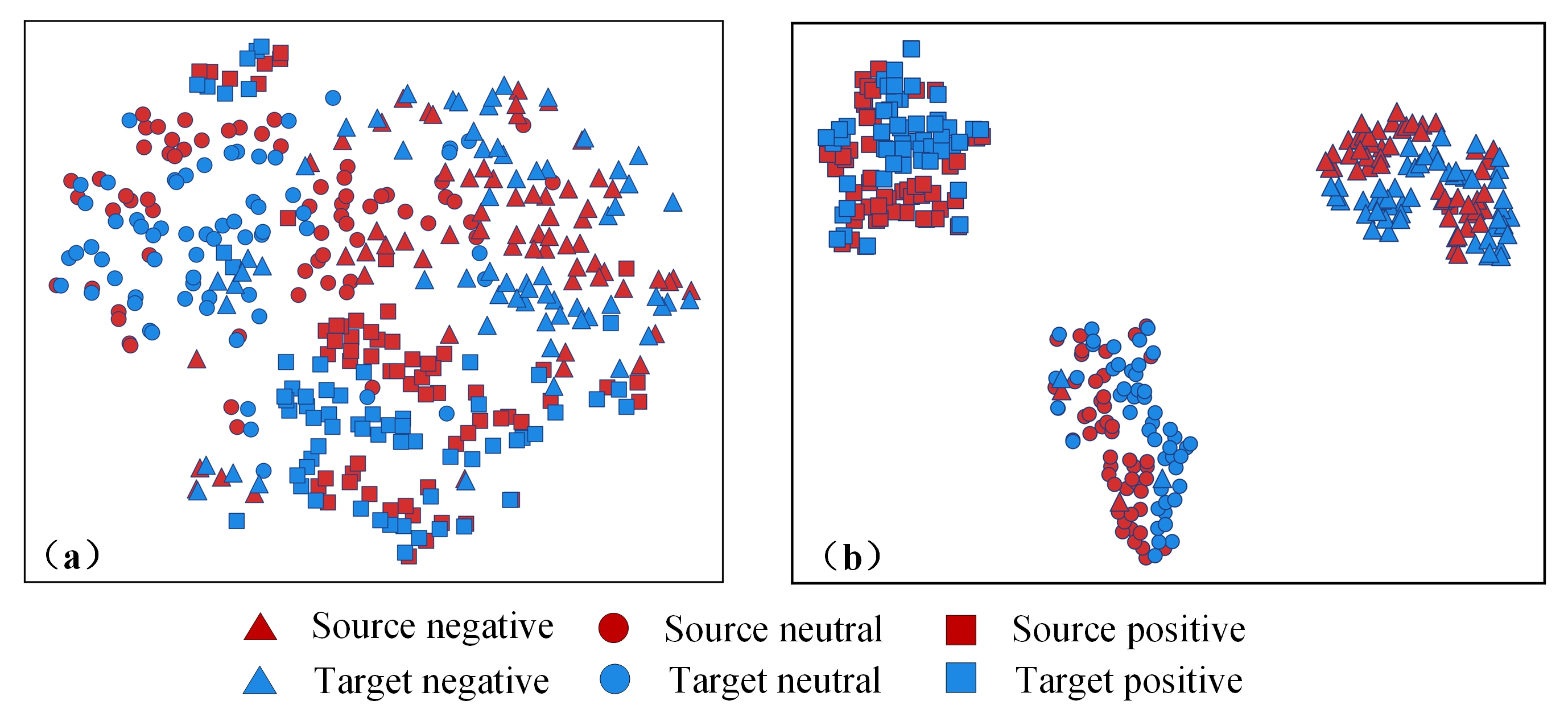}
\caption{t-SNE visualization of feature distributions: (a) before training; (b) after training.}
\label{fig_7}
\end{figure}
\begin{figure}[t!]\centering
\centering
\includegraphics[width=3.4in]{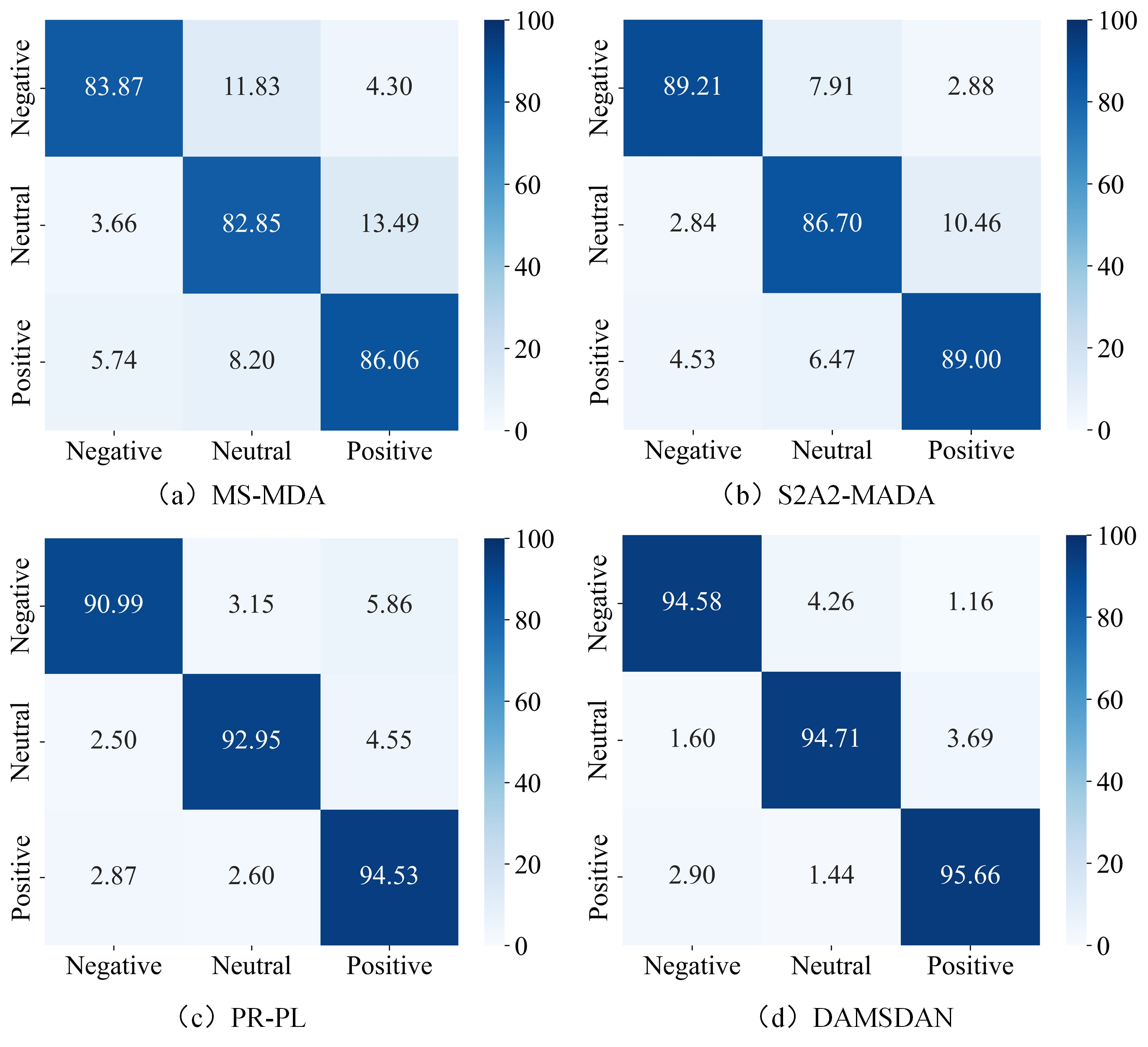}
\caption{Confusion matrices of MS-MDA, S2A2-MADA, PR-PL, and DAMSDAN on the SEED dataset.}
\label{fig_8}
\end{figure}
\subsection{t-SNE}
To further assess the effectiveness of DAMSDAN in achieving both domain alignment and discriminative feature learning, we employ the t-SNE method~\cite{ref44} to visualize the high-dimensional feature representations on the SEED dataset, as depicted in Fig.~\ref{fig_7}. In Fig.~\ref{fig_7}(a), prior to training, source (red) and target (blue) samples are extensively intermingled in the feature space, reflecting poor domain separability. Moreover, the emotional categories (negative, neutral, and positive, represented by squares, triangles, and circles, respectively) exhibit significant interclass overlap, characterized by dispersed intraclass distributions and indistinct boundaries. In contrast, Fig.~\ref{fig_7}(b) displays the feature distribution after training with DAMSDAN. Compared with the original representation, the trained model yields significantly improved domain alignment and clearer semantic structure. Samples from different domains but belonging to the same emotional category are more tightly clustered, indicating enhanced intraclass compactness and improved interclass separability. Although a few samples remain close to decision boundaries, most are clearly clustered within their respective semantic regions. These findings confirm DAMSDAN's capability to learn representations that are both emotion-discriminative and domain-invariant, thereby validating its robustness and generalization performance in cross-domain EEG-based emotion recognition.

\subsection{Confusion Matrix}
To further evaluate the discriminative capability of DAMSDAN in EEG-based emotion classification tasks, we present the confusion matrix on the SEED dataset and perform a comparative analysis with several representative baseline models, as illustrated in Fig.~\ref{fig_8}. As shown in Fig.~\ref{fig_8}(d), DAMSDAN achieves classification accuracies exceeding 94\% across all three emotional categories, demonstrating both strong and consistent performance. Specifically, the DAMSDAN model attains accuracies of 95.66\%, 94.71\%, and 94.58\% for the positive, neutral, and negative classes, respectively, with corresponding misclassification rates constrained within the range of 1.16\% to 4.26\%. In comparison, the compared methods in Fig.~\ref{fig_8}(a)-(c) exhibit notably inferior performance across all emotional categories. Compared with the second-best model, PR-PL, DAMSDAN yields absolute accuracy gains of 1.13\%, 1.76\%, and 3.59\% for the positive, neutral, and negative categories, respectively. The performance gap widens further when compared with the weakest model, MS-MDA, showing improvements of 9.6\%, 11.86\%, and 10.71\% for the positive, neutral, and negative categories, respectively. These results underscore the effectiveness of DAMSDAN in reducing inter-class confusion and maintaining balanced performance across emotional states, thereby validating its robustness and generalization capability in cross-domain EEG-based emotion recognition.
\begin{figure}[t!]\centering
\centering
\includegraphics[width=3.4in]{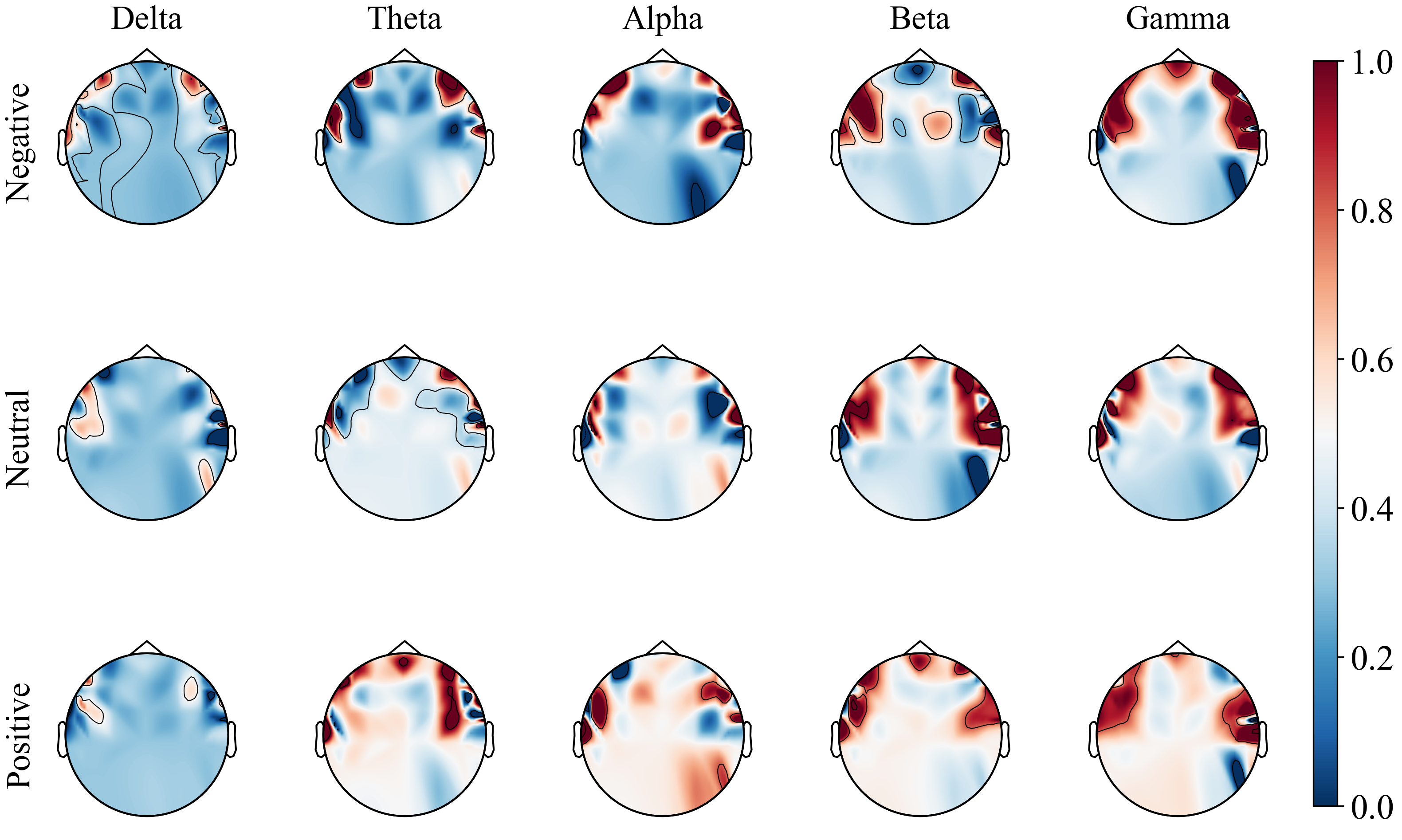}
\caption{Topographical analysis of mutual information between EEG patterns and model predictions.}
\label{fig_9}
\end{figure}

\subsection{Topographical Analysis of Salient EEG Patterns}
During the topological feature analysis stage, mutual information is computed between EEG input features and model prediction outputs on the SEED dataset to identify the most informative frequency bands and brain regions contributing to emotion recognition. Specifically, in the $j$-th validation fold, the target domain input samples are denoted as $x_{j}^{T}\in \mathbb{R}^{N_T\times D_T}$, where ${N_T}$ represents the number of target samples and ${D_T}$ denotes the total dimensionality of DE features extracted from five frequency bands across 62 EEG channels. The corresponding model output is denoted as $\hat{y}_{j}^{T}\in \mathbb{R}^{N_T\times 3}$, representing the predicted probability distributions over the three emotional classes: negative, neutral, and positive. To quantify the contribution of different EEG patterns to emotion classification, a non-parametric mutual information estimation approach, as proposed in~\cite{ref45, ref46, ref47}, is employed to compute the mutual information matrix ${\bf{I}}\left( {x_j^T,\hat y_j^T} \right) \in {\mathbb{R}^{3 \times {D_T}}}$ between $x_j^T$ and $\hat y_j^T$. This matrix characterizes the statistical dependence between individual frequency-channel feature pairs and the predicted emotional outputs. To ensure consistency across validation folds, the computed mutual information values are normalized to the range $\left[ {0,1} \right]$, where higher values indicate greater discriminative contributions from the corresponding feature modality. Fig.~\ref{fig_9} illustrates the average normalized mutual information values across all validation folds. The results reveal that among all frequency bands, the $\beta $ and $\gamma $ bands carry the most salient information. Furthermore, the corresponding discriminative features are primarily concentrated in the prefrontal and temporal regions, which are critically involved in affective processing and contribute substantially to robust cross-domain emotion recognition.

\section{Conclusion}
\label{section6}
In this paper, we introduce a novel deep learning-driven domain adaptation approach, termed DAMSDAN, specifically designed to address the challenges of individual variability and distributional shifts in cross domain EEG-based emotion recognition. The proposed framework is organized into three core modules: FE module, MDA module, and CDA module. These components collectively aim to improve cross-domain generalization and emotional class discriminability. In the feature encoding stage, DAMSDAN integrates both common and domain-specific encoders to effectively extract emotion relevant and transferable EEG representations. For marginal distribution alignment, DAMSDAN integrates the PCC mechanism with the ADA mechanism to simultaneously improve intra-class compactness and inter-class separability. In addition, a DASW strategy is introduced to dynamically adjust the contribution of each source domain based on its distributional relevance to the target domain, effectively suppressing negative transfer. For conditional distribution alignment, DAMSDAN integrates a DPLC strategy together with a PGCA strategy to enhance the stability of pseudo-label supervision and facilitate fine-grained class-level alignment. Extensive experiments conducted on three publicly available EEG emotion datasets, namely SEED, SEED IV, and FACED, demonstrate that DAMSDAN outperforms state-of-the-art methods in both cross-subject and cross-session scenarios. In addition, ablation studies further validate the effectiveness and indispensability of each proposed module. In conclusion, DAMSDAN offers a robust and scalable solution for cross-domain EEG-based emotion recognition and establishes a solid foundation for future research in domain-adaptive affective brain-computer interface systems.



\vfill

\begin{thebibliography}{99}
\bibitem{ref1} 
H. Huang, et al., ``An EEG-based brain computer interface for emotion recognition and its application in patients with disorder of consciousness,'' {\it{IEEE Transactions on Affective Computing}}, vol. 12, no. 4, pp. 832--842, 2021.

\bibitem{ref2} 
Y. Bian, C. Zhou, Y. Zhang, J. Liu, J. Sheng and Y.-J. Liu, ``Focus on Cooperation: A Face-to-Face VR Serious Game for Relationship Enhancement,'' {\it{IEEE Transactions on Affective Computing}}, vol. 15, no. 3, pp. 913--928, 2024.

\bibitem{ref3} 
K. Magtibay and K. Umapathy, ``A Review of Tools and Methods for Detection, Analysis, and Prediction of Allostatic Load Due to Workplace Stress,'' {\it{IEEE Transactions on Affective Computing}}, vol. 15, no. 1, pp. 357--375, 2024.

\bibitem{ref4} 
D. Wu, X. Jiang and R. Peng, ``Transfer learning for motor imagery based brain--computer interfaces: A tutorial,'' {\it{Neural Networks}}, vol. 153, pp. 235--253, 2022.

\bibitem{ref5} 
H. Li, Y.-M. Jin, W.-L. Zheng, and B.-L. Lu, ``Cross-subject emotion recognition using deep adaptation networks,'' in {\it{Proc. 25th Int. Conf. Neural Inf. Process. Systems (NeurIPS)}}, Siem Reap, Cambodia, pp. 403--413, 2018.

\bibitem{ref6} Y. Ma, W. Zhao, M. Meng, Q. Zhang, Q. She and J. Zhang, ``Cross-Subject Emotion Recognition Based on Domain Similarity of EEG Signal Transfer Learning,'' {\it{IEEE Transactions on Neural Systems and Rehabilitation Engineering}}, vol. 31, pp. 936--943, 2023.

\bibitem{ref7} 
Y. Li, W. Zheng, Y. Zong, Z. Cui, T. Zhang and X. Zhou, ``A bi-hemisphere domain adversarial neural network model for EEG emotion recognition,'' {\it{IEEE Transactions on Affective Computing}}, vol. 12, no. 2, pp. 494--504, 2021.

\bibitem{ref8} 
R. Zhou, et al., ``PR-PL: A Novel Prototypical Representation Based Pairwise Learning Framework for Emotion Recognition Using EEG Signals,'' {\it{IEEE Transactions on Affective Computing}}, vol. 15, no. 2, pp. 657--670, 2024.

\bibitem{ref9} 
Y. Yang et al., ``Spectral-Spatial Attention Alignment for Multi-Source Domain Adaptation in EEG-Based Emotion Recognition,'' {\it{IEEE Transactions on Affective Computing}}, vol. 15, no. 4, 2024.

\bibitem{ref10} J. Cao, X. He, C. Yang, S. Chen, Z. Li, and Z. Wang, ``Multi-source and multi-representation adaptation for cross-domain electroencephalography emotion recognition,'' {\it{Frontiers in Psychology}}, vol. 12, no. 6516, 2022.

\bibitem{ref11} 
H. Chen, M. Jin, Z. Li, C. Fan, J. Li, and H. He, ``MS-MDA: Multisource marginal distribution adaptation for cross-subject and cross-session EEG emotion recognition,'' {\it{Frontiers in Neuroscience}}, vol. 15, no. 778488, 2021.

\bibitem{ref12} 
J. Li, S. Qiu, Y.-Y. Shen, C.-L. Liu, and H. He, ``Multisource transfer learning for cross-subject EEG emotion recognition,'' {\it{IEEE Transactions on Cybernetics}}, vol. 50, no. 7, pp. 3281--3293, 2019.

\bibitem{ref13} 
M. Long, et al., ``Conditional adversarial domain adaptation,'' {\it{Advances in Neural Information Processing Systems}}, vol. 31, 2018.

\bibitem{ref14} 
Y. Ye, X. Zhu, Y. Li, T. Pan and W. He, ``Cross-subject EEG-based Emotion Recognition Using Adversarial Domain Adaption with Attention Mechanism,'' in {\it{Proc. 43rd Annu. Int. Conf. IEEE Eng. Med. Biol. Soc. (EMBC)}}, Mexico City, Mexico, pp. 1140--1144, 2021.

\bibitem{ref15} 
L. Zhang, J. Fu, S. Wang, D. Zhang, Z. Dong, and C. P. Chen, ``Guide subspace learning for unsupervised domain adaptation,'' {\it{IEEE Transactions on Neural Networks and Learning Systems}}, vol. 31, no. 9, pp. 3374--3388, 2020.

\bibitem{ref31} 
Z. Li, et al., ``Dynamic domain adaptation for class-aware cross-subject and cross-session EEG emotion recognition,'' {\it{IEEE Journal of Biomedical and Health Informatics}}, vol. 26, no. 12, pp. 5964--5973, 2022.

\bibitem{ref43} 
X. Hong, C. Du and H. He, ``Adaptive Domain Alignment Neural Networks for Cross-Domain EEG Emotion Recognition,'' {\it{IEEE Transactions on Affective Computing}}, vol. 16, no. 2, pp. 903--914, 2024. 

\bibitem{ref18} M. Soleymani, S. Asghari-Esfeden, Y. Fu, and M. Pantic, ``Analysis of EEG signals and facial expressions for continuous emotion detection,'' {\it{IEEE Transactions on Affective Computing}}, vol. 7, pp. 17--28, 2015.

\bibitem{ref19} 
H. Zhao, R. T. des Combes, K. Zhang, and G. J. Gordon, ``On learning invariant representation for domain adaptation,'' in {\it{Proc. Int. Conf. Machine Learning (ICML)}}, Long Beach, CA, USA, pp. 7523--7532, 2019.

\bibitem{ref20} 
W.-L. Zheng and B.-L. Lu, ``Investigating Critical Frequency Bands and Channels for EEG-based Emotion Recognition with Deep Neural Networks,'' {\it{IEEE Transactions on Autonomous Mental Development}}, vol. 7, no. 3, pp. 162--175, 2015.

\bibitem{ref21} 
W.-L. Zheng, W. Liu, Y. Lu, B.-L. Lu and A. Cichocki, ``EmotionMeter: A Multimodal Framework for Recognizing Human Emotions,'' {\it{IEEE Transactions on Cybernetics}}, vol. 49, no. 3, pp. 1110--1122, 2019.

\bibitem{ref22} 
J. Chen, X. Wang, C. Huang, X. Hu, X. Shen, and D. Zhang, ``A large finer-grained affective computing EEG dataset,'' {\it{Scientific Data}}, vol. 10, no. 1,  pp. 740, 2023.

\bibitem{ref38} 
H. Zhang, T. Zuo, Z. Chen, X. Wang and P. Z. H. Sun, ``Evolutionary Ensemble Learning for EEG-Based Cross-Subject Emotion Recognition,'' {\it{IEEE Journal of Biomedical and Health Informatics}}, vol. 28, no. 7, pp. 3872--3881, 2024. 

\bibitem{ref16} 
Y. Zhou et al., ``Cross-Subject Cognitive Workload Recognition Based on EEG and Deep Domain Adaptation,'' {\it{IEEE Transactions on Instrumentation and Measurement}}, vol. 72, no. 2518912, pp. 1--12, 2023.

\bibitem{ref17} 
Y. Wang, J. Wang, W. Wang, J. Su, C. Bunterngchit and Z.-G. Hou, ``TFTL: A Task-Free Transfer Learning Strategy for EEG-Based Cross-Subject and Cross-Dataset Motor Imagery BCI,'' {\it{IEEE Transactions on Biomedical Engineering}}, vol. 72, no. 2, pp. 810--821, 2025.

\bibitem{ref23} 
Y. Wang, L. Zhang and Y. Zhang, ``Band-Level Adaptive Fusion Network for Cross-Subject EEG Emotion Recognition,'' {\it{IEEE Transactions on Instrumentation and Measurement}}, vol. 74, pp. 1--12, no. 2514112, 2025.

\bibitem{ref27} 
M. Long, J. Wang, G. Ding, J. Sun and P. S. Yu, ``Transfer feature learning with joint distribution adaptation,'' in {\it{Proc. IEEE Int. Conf. Computer Vision (ICCV)}}, Sydney, Australia, pp. 2200--2207, 2013.

\bibitem{ref24} 
Z. Lan, O. Sourina, L. Wang, R. Scherer and G. R. Müller-Putz, ``Domain adaptation techniques for EEG-based emotion recognition: A comparative study on two public datasets,'' {\it{IEEE Transactions on Cognitive and Developmental Systems}}, vol. 11, no. 1, pp. 85--94, 2019.

\bibitem{ref25} 
S. J. Pan, I. W. Tsang, J. T. Kwok and Q. Yang, ``Domain adaptation via transfer component analysis,'' {\it{IEEE Transactions on Neural Networks}}, vol. 22, no. 2, pp. 199--210, 2011.

\bibitem{ref26} 
L. B.-Q. Ma, H. Li, W.-L. Zheng, and B.-L. Lu, ``Reducing the subject variability of EEG signals with adversarial domain generalization,'' in {\it{Proc. Int. Conf. Neural Information Processing (ICONIP)}}, Cham, Switzerland, pp. 30--42, 2019.

\bibitem{ref28} 
J. Wang, W. Feng, Y. Chen, H. Yu, M. Huang and P. S. Yu, ``Visual domain adaptation with manifold embedded distribution alignment,'' in {\it{Proc. 26th ACM Multimedia Conf. (ACM MM)}}, Seoul, South Korea, pp. 402--410, 2018.

\bibitem{ref29} 
L. Zhu, et al., ``Multisource Wasserstein adaptation coding network for EEG emotion recognition,'' {\it{Biomedical Signal Processing and Control}}, vol. 76, 2022.

\bibitem{ref30} 
N. She, C. Zhang, F. Fang, Y. Ma and Y. Zhang, ``Multisource Associate Domain Adaptation for Cross-Subject and Cross-Session EEG Emotion Recognition,'' {\it{IEEE Transactions on Instrumentation and Measurement}}, vol. 72, pp. 1--12, 2023.

\bibitem{ref32} 
J. Li, S. Qiu, C. Du, Y. Wang, and H. He, ``Domain adaptation for EEG emotion recognition based on latent representation similarity,'' {\it{IEEE Transactions on Cognitive and Developmental Systems}}, vol. 12, no. 2, pp. 344--353, 2020.

\bibitem{ref33} 
R. Zhang, H. Guo, Z. Xu, Y. Hu, M. Chen, and L. Zhang, ``MGFKD: A semi-supervised multi-source domain adaptation algorithm for cross-subject EEG emotion recognition,'' {\it{Brain Research Bulletin}}, vol. 208, no. 110901, 2024.

\bibitem{ref34} 
Y. Yang et al., ``Spectral-Spatial Attention Alignment for Multi-Source Domain Adaptation in EEG-Based Emotion Recognition,'' {\it{IEEE Transactions on Affective Computing}}, vol. 15, no. 4, pp. 2012--2024, 2024.

\bibitem{ref35} 
K. Liu et al., ``Enhancing EEG-Based Cross-Subject Emotion Recognition via Adaptive Source Joint Domain Adaptation,'' {\it{IEEE Transactions on Affective Computing}}, 2024.

\bibitem{ref36} 
T. Song, W. Zheng, P. Song and Z. Cui, ``EEG emotion recognition using dynamical graph convolutional neural networks,'' {\it{IEEE Transactions on Affective Computing}}, vol. 11, no. 3, pp. 532--541, 2020.

\bibitem{ref37} 
W. Liu, J.-L. Qiu, W.-L. Zheng and B.-L. Lu, ``Comparing recognition performance and robustness of multimodal deep learning models for multimodal emotion recognition,'' {\it{IEEE Transactions on Cognitive and Developmental Systems}}, vol. 14, no. 2, pp. 715--729, 2022.


\bibitem{ref39} 
H. Chen, Z. Li, M. Jin, and J. Li, ``MEERNet: Multi-source EEG-based emotion recognition network for generalization across subjects and sessions,'' in {\it{Proc. IEEE 43rd Annu. Int. Conf. Eng. Med. Biol. Soc.}}, pp. 6094--6097, 2021.

\bibitem{ref40} 
J. Chang, Z. Zhang, Y. Qian and P. Lin, ``Multi-Scale Hyperbolic Contrastive Learning for Cross-Subject EEG Emotion Recognition,'' {\it{IEEE Transactions on Affective Computing}}, 2025.

\bibitem{ref41} 
X. Shen, X. Liu, and Z. S. Song, ``Contrastive learning of subject invariant EEG representations for cross-subject emotion recognition,'' {\it{IEEE Transactions on Affective Computing}}, vol. 14, no. 3, pp. 2496--2511, 2023.

\bibitem{ref42} 
J. Cao, X. He, C. Yang, S. Chen, Z. Li, and Z. Wang, ``Multisource and multi-representation adaptation for cross-domain electroencephalography emotion recognition,'' {\it{Frontiers in Psychology}}, vol. 12, no. 6516, 2022. 

\bibitem{ref44} 
L. van der Maaten and G. Hinton, ``Visualizing data using t-SNE,'' {\it{Journal of Machine Learning Research}}, vol. 9, pp. 2579--2605, 2008.

\bibitem{ref45} 
L. Kozachenko and N. Leonenko, ``Sample estimate of the entropy of a random vector,'' {\it{Problemy Peredachi Informatsii}}, vol. 23, no. 2, pp. 9--16, 1987.

\bibitem{ref46} 
A. Kraskov, H. Stögbauer, and P. Grassberger, ``Estimating mutual information,'' {\it{Physical review E}}, vol. 69, no. 6, pp. 066138, 2004.

\bibitem{ref47} 
B. Ross, ``Mutual information between discrete and continuous data sets,'' {\it{Plos One}}, vol. 9, no. 2, pp. e87357, 2014.
\end{thebibliography}
\end{document}